\documentclass[final,5p,times,twocolumn]{elsarticle}

\pdfoutput=1

\usepackage{array}  

\usepackage{amsmath}

\usepackage{float}

\newcolumntype{C}[1]{>{\centering\arraybackslash}p{#1}}

\usepackage{caption}
\usepackage{graphicx}

\usepackage{tabularx}

\usepackage{booktabs} 

\usepackage{multirow} 
\usepackage{array}    

\usepackage{bm}

\usepackage{amssymb}

\begin{document}

\begin{frontmatter}



\title{An Augmented Surprise-guided Sequential Learning Framework for Predicting the Melt Pool Geometry}


\author[inst1]{Ahmed Shoyeb Raihan}

\affiliation[inst1]{organization={Department of Industrial and Management Systems Engineering, West Virginia University},
            city={Morgantown},
            postcode={26506}, 
            state={West Virginia},
            country={USA}}
\affiliation[inst2]{organization={Department of Mechanical Engineering, The University of Texas at San Antonio},
            city={San Antonio},
            postcode={78249}, 
            state={Texas},
            country={USA}}

\author[inst1]{Hamed Khosravi}

\author[inst2]{Tanveer Hossain Bhuiyan}

\author[inst1]{Imtiaz Ahmed}


\begin{abstract}
Metal Additive Manufacturing (MAM) has transformed the manufacturing landscape, bringing notable benefits such as intricate design capabilities, minimal material wastage, rapid prototyping, compatibility with diverse materials, and customized solutions. However, the complete adoption of this technology in the industry is impeded by challenges in ensuring uniform product quality. A pivotal aspect of MAM's successful implementation lies in understanding the intricate relationship between its process parameters and the melt pool characteristics. In this scenario, the integration of Artificial Intelligence (AI) into MAM is vital. While traditional machine learning (ML) approaches are effective, they typically rely on large datasets to accurately capture complex relationships. However, in MAM, creating such extensive datasets is both time-consuming and resource-intensive, posing a significant challenge for effectively applying these methods. Our study addresses this challenge by introducing a novel surprise-guided sequential learning framework (SurpriseAF-BO). The framework represents a paradigm shift in the field of MAM, leveraging an iterative and adaptive learning approach. It efficiently models the dynamics between process parameters and melt pool characteristics with limited data which is a critical advantage in the cyber manufacturing environment of MAM. In comparison to conventional machine learning models, our sequential learning method demonstrates superior predictive accuracy for melt pool depth, width, and length. To further improve our methodology, we have incorporated a Conditional Tabular Generative Adversarial Network (CTGAN) model into our framework. It produces synthetic data that closely mirrors real experimental data, enhancing the overall learning process. We call this improved framework `CT-SurpriseAF-BO'. This advancement in our sequential learning approach significantly strengthens its predictive accuracy, all while eliminating the need for extra physical experiments. Our study not only showcases the application of cutting-edge data-driven techniques but also highlights the significant impact of sequential AI and ML in the field of cyber manufacturing, especially in MAM.
\end{abstract}




\end{frontmatter}

\makeatletter
\def\ps@pprintTitle{%
 \let\@oddhead\@empty
 \let\@evenhead\@empty
 \def\@oddfoot{\reset@font\hfil\thepage\hfil}
 \let\@evenfoot\@oddfoot
}
\makeatother


\section{Introduction}
\label{sec:sample1}
With capabilities such as design flexibility, high customizability, and the ability to produce parts with complex structures, metal additive manufacturing (MAM) has emerged as a revolutionary manufacturing method in recent years \cite{berman20123,ngo2018additive}. Compared to conventional manufacturing techniques, MAM has various benefits, such as a shorter lead time, less material usage, and improved performance and reliability, which have led to its widespread adoption in the aerospace, automotive, and biomedical industries \cite{debroy2018additive,jiang2020path,leal2017additive,liu2021review,wang2016topological}. Nevertheless, while these advantages have spurred new research interests and applications in both academic and industrial sectors, there are critical challenges that need to be tackled, such as processing defects and variable product quality \cite{wang2020machine}. Specifically, the quality of printed components is a major concern, as defects can undermine the structural integrity of the parts. This issue is particularly challenging due to the complex, multiscale physics inherent in the additive manufacturing (AM) process, making it one of the most difficult problems to solve \cite{akbari2022meltpoolnet,seet2016multi}. 

In this regard, different experimental monitoring methods, mainly categorized as ex-situ and in-situ methods, are commonly used to identify the ideal processing conditions for producing additively manufactured parts with minimal quality issues. Ex-situ techniques analyze printed parts after production, while in-situ techniques monitor the manufacturing process in real-time. By employing these monitoring techniques manufacturers can gain a deeper understanding of how defects are formed and improve their defect mitigation strategies \cite{bayat2021review}. Particularly, in-situ monitoring of the melt pool is critical for ensuring part quality due to the fact that most of the common defects in MAM, i.e., lack of fusion, keyhole, porosity, and balling, are rooted in the dynamics of the melt pool \cite{akbari2022meltpoolnet,scime2019using}. Monitoring and understanding the behavior of the melt pool, which refers to the localized area where metal powder is selectively melted and solidified to create the desired part or component, are critical for process optimization. By analyzing the melt pool characteristics, such as dimensions, shape, and energy input, process parameters can be adjusted to achieve desired outcomes, such as minimizing residual stresses, reducing distortion, and enhancing part quality. The quality of the final product is directly connected to the characteristics of the melt pool \cite{ye2023predictions}. Therefore, it is important to achieve desired melt pool characteristics with certain a geometry (depth, width, and length), which is dependent on different process parameters such as beam power, scanning speed, beam diameter, hatch spacing, and so on \cite{gordon2020defect}. The nature of this relationship is often complex and unknown to the manufacturing engineers. It can only be unveiled through physical experiments which are typically expensive. In the literature this is termed as a black-box function, which is difficult to optimize or approximate. Historically, researchers have used different physical and mechanical models to relate the process parameters with the melt pool geometry. However, because of the complex multi-physics and multi-scale characteristics of AM processes and the dominant influence of processing parameters on the quality of printed products, research in the field of AM has, in recent years, moved away from physics-based approaches towards data-driven approaches \cite{wang2020machine}. Consequently, data-driven analysis and ML have become commonplace in AM applications and are gaining traction in recent research related to AM \cite{johnson2020invited}. With a reliable training dataset consisting of process parameters as input features and melt pool depth, width and length as labels, the trained ML models can make accurate predictions, and determine the optimal processing parameters to achieve a desired melt pool dimension in an efficient manner. 

Despite their widespread application, traditional ML algorithms require large amounts of experimental data initially for training the models, and these data can only be obtained from physical experiments \cite{johnson2020invited}. The prediction accuracy largely depends on the availability of previously conducted experiments. However, experimental data are not always available beforehand, and it is an expensive and time-consuming process to obtain data by performing physical experiments \cite{khosravi2023data}. Moreover, as most of these ML techniques involve a batch learning process, large datasets, which need to be available upfront, are fed into the training block of these algorithms for the learning purpose. This limitation further restricts their ability to adapt to evolving data distributions \cite{ramezankhani2021making}. Furthermore, these ML methods with batch learning mechanisms are not very effective in utilizing the available data since they treat all samples equally, potentially wasting computational resources on less informative or redundant samples.

Because of these limitations, very recently, researchers have recognized the significance of sequential learning, particularly Bayesian optimization (BO), acknowledging that a comprehensive understanding of a complex system cannot be achieved solely through a single instance of action \cite{ahmed2021towards}. In BO, a surrogate model and an acquisition function works hand in hand to discover the black-box function. In contrast to traditional ML models, here, surrogate models can adapt themselves based on the outcomes of physical experiments. On the other hand, at each iteration, the acquisition function selects the most promising location to update surrogate models’ understanding on the underlying black-box function, i.e., the relationship between the process parameters and the melt pool characteristics. It is more intelligent, cognizant and economical compared to the static ML models which are data hungry and cannot incorporate real-time knowledge to guide their search process. Sequential learning models, although more efficient than static ML models, have their own caveats. The problem lies in the acquisition strategy embedded in these algorithms. While selecting the next experiment location (i.e., combination of process parameters), these algorithms balance two actions, typically known as exploration and exploitation. While exploration helps the algorithm to escape from settling in a local optimum region and thereby potentially miss a part of the functional space, its counterpart exploitation helps the algorithm to exploit its current knowledge and stay close to the familiar landscape of the functional space. It is important to balance these two actions to approximate the black-box function in fewer iterations. The two most widely used acquisition functions in the BO literature are expected improvement (EI) and upper confidence bound (UCB) \cite{wilson2018maximizing}. While the former favors exploitation, the latter favors exploration and thus makes them fall into two extremes. 

To overcome these challenges, we propose a novel acquisition function based on ‘surprise’ under the BO framework and thus a modified surprise-guided sequential learning framework known as ‘SurpriseAF-BO’. Surprise is an unusual finding from an experiment that does not match the current understanding of the system. We believe that the experimental search strategy should mimic the thought process of a human scientist and its change of course should be directed by surprise only. Therefore, in our proposed framework, the balance between exploration and exploitation is achieved by this new surprise-guided acquisition strategy. In our earlier works, the surprise guided framework shows a promising result compared to the other sequential approaches \cite{ahmed2021towards, jin2022autonomous}.

The BO framework generally begins with a set of initial experiments aimed at priming the surrogate model. Following this `warm-up' phase, it shifts into a sequential mode, where it systematically selects further experiments to approximate the underlying relationship more accurately.  During our experimentation, we realized that having more initial experiments can only help the learning process. As real experiments are expensive and time consuming, we utilize a Conditional Tabular Generative Adversarial Network (CTGAN) to augment the initial training process by generating synthetic experiments. This leads to our final model which is termed as `CT-SurpiseAF-BO'. As per our expectation, we find that the CTGAN generated samples when combined with the original samples further enhance the performance of the proposed framework. 

For comparison purposes, we divide our evaluation experiments into two phases. In the first phase (Phase I), to emphasize the importance of sequential experiment design, we show that, in predicting the geometry of the melt pool, our proposed surprise-guided active learning framework (SurpriseAF-BO) performs better than the regularly used ML algorithms in a data-scarce environment. The results have been compared with six popular ML algorithms: Radom Forest (RF), Support Vector Regression (SVR), Gradient Boosting (GB) and Neural Network (NN). In the second phase (Phase II), we introduce an integrated model where we combine surprise-guided framework with CTGAN. We see an improved performance in the RMSE metric in all the three cases of melt pool geometry (depth, width, and length) prediction with this extended proposed framework (CT-SurpiseAF-BO).

The main contributions of this work, therefore, can be summarized as:

\begin{itemize}
    \item We propose a surprise-guided acquisition function in a formal sequential learning framework (SurpriseAF-BO) to approximate a black-box function.
    \item We extend this proposed surprise-guided sequential learning framework by introducing CT-SurpiseAF-BO, which integrates a CTGAN model to further improve the performance of our proposed approach under a resource-constrained environment.
    \item We use both the original SurpriseAF-BO and augmented CT-SurpiseAF-BO frameworks in approximating the relationship between the melt pool geometry (depth, width, and length) and process parameters utilizing a melt pool dataset.
    \item We compare their performance with the EI-based BO framework and six classical ML techniques.
\end{itemize}

The rest of the paper is designed as follows: In Section 2, we discuss some relevant works in the literature regarding melt pool characterization in metal additive manufacturing using model-based, data-driven, and ML methods. We also discuss BO, one of the most widely used sequential frameworks, with its applications and limitations. The methodology of our proposed sequential learning frameworks is introduced in Section 3. In Section 4, we provide an elaborate discussion of the melt pool dataset and data processing steps. The hyperparameters of the used ML methods are also explained in this section. Section 5 presents the findings of this study from the two phases. In Phase I, we use a melt pool dataset to evaluate the performance of our proposed approach (SurpriseAF-BO) and compare the results with the traditional EI-based BO and six popular ML techniques that are widely used in MAM applications. In Phase II, we use the same melt pool dataset, but this time augmenting the surprise-guided framework with CTGAN. We illustrate the improved performance of our enhanced CT-SurpiseAF-BO approach. Lastly, we discuss our findings and conclude the work in Section 6.

\section{Literature Review}
\label{sec:sample2}

MAM offers significant advantages in the manufacturing industry. It allows for the production of complex geometries and provides design flexibility, enabling the creation of intricate and optimized parts \cite{bandyopadhyay2020recent}. The process reduces material waste, leading to cost savings, and enables the consolidation of multiple components into a single part, simplifying assembly and reducing failure risks \cite{ford2016additive}. In addition, MAM facilitates rapid prototyping and iteration, speeding up the product development cycle. It supports a wide range of materials, providing versatility for various applications \cite{madhavadas2022review}. Moreover, it reduces lead times, enables on-demand manufacturing, and allows for customization and personalization of parts, catering to individual customer needs \cite{guo2022machine}. However, despite all these advantages, MAM still suffers from issues related to the quality and consistency of the final product. 

It is important to understand the mechanisms and reasons behind the generation of defects in additively manufactured metal parts \cite{debroy2018additive}. Gaining insight into the defects formed in MAM requires a fundamental understanding of how the process parameters impact the characteristics of the melt pool \cite{matthews2020controlling,mondal2020investigation}. The melt pool, which is the liquid interface between the powder particles and the energy source during the AM process, is the region where defects such as lack of fusion, keyholing, balling, and beading occur. As a result, to achieve products with high quality consistently, the properties of the melt pool should be controlled which in turn is correlated to the process parameters chosen in MAM. Process parameters such as laser beam power, scanning velocity, beam diameter, hatch spacing, and melting temperature have significant influence in the melt pool properties such as dimensions or geometry (length, depth, and width) \cite{gordon2020defect}. Since achieving the desired part quality without defects lies in the control of melt pool properties, it is mandatory to understand the latent relation between these process parameters and melt pool properties \cite{akbari2022meltpoolnet}. Consequently, there have been several works in the existing literature that aim to learn about melt pool dynamics to gain more insights that could be useful to manufacture defect free 3D printed parts. While earlier works make use of simulations, analytical and numerical models to understand the melt pool properties, recent studies focus on data-driven approaches. 

Several model-based studies have been proposed over the years. Investigations into the behavior of melt pools in the selective laser melting (SLM) process have led to the development of a 3D simulation exploring melt pool dynamics in a powder bed with a random particle distribution, employing the discrete element method (DEM) for realistic simulation conditions \cite{wu2018numerical}. The neighboring-effect modeling method (NBEM) has been introduced for predicting melt pool size in additive manufacturing, utilizing physical principles of melt pool formation and scan strategies for optimization and real-time process control \cite{yang2020scan}. A combination of numerical analysis and experimental data has been used to comprehensively investigate temperature distribution and melt pool size in the SLM process, employing a 3D finite element analysis with a moving volumetric Gaussian laser heat source \cite{ansari2019investigation}. Further studies have developed a 3D model to predict the thermal behavior and geometry of melt pools, focusing on laser material interaction and examining both stationary and moving laser beam scenarios \cite{Han2005ThermalBA}. The impact of hatch spacing on the microstructure and melt pool properties of 316L stainless steel has been analyzed, revealing that increased hatch spacing leads to deeper but narrower melt pools, as shown through three-dimensional Finite Element Method (FEM) simulations \cite{dong2018effect}. Additionally, a novel hybrid heat source model has been introduced to predict various melt pool characteristics, including dimensions and melting modes, within the SLM process, taking into account different absorption mechanisms for materials in both porous and dense states \cite{lee2020novel}. A three-phase model employing the Volume of Fluid (VOF) method has been proposed to investigate heat transfer and melt pool behavior in the LPBF process, integrating factors like surface tension, the Marangoni effect, and recoil pressure, with results indicating that laser power and scanning speed are significant influencers of melt pool dimensions and shape \cite{li2021three}. Lastly, a 3D numerical model for SLM has been constructed, incorporating various physical effects to enable a detailed understanding of SLM melt pool dynamics with relatively low computational effort \cite{heeling2017melt}.

These model-based approaches, however, suffers from some major drawbacks \cite{zhang2021prediction}. Analytical modeling is limited by assumptions that might not accurately reflect real-world complexities. In the context of melt pool properties prediction, the volume and mass changes of the part over time and uncertainties in the thermo-physical process can make analytical solutions unreliable \cite{khanzadeh2019situ,shamsaei2015overview}. Additionally, numerical modeling, particularly finite element methods (FEM), requires detailed information and parameters that might not be fully available, such as heat loss caused by powder and substrate reflection \cite{xiong2009situ}. Moreover, the precision of FEM is significantly influenced by factors such as element types, boundary conditions, constitutive models, and meshing schemes, potentially leading to the introduction of uncertainties \cite{luo2018survey}. In this regard, data-driven approaches, like ML, can help overcome the limitations of analytical and numerical modeling in melt pool properties prediction. These approaches offer the benefit of accurately and efficiently modeling complex physical problems, even when there is only a modest understanding of the underlying physics of the problem \cite{wuest2016machine}. By leveraging large datasets of experimental and simulated melt pool behavior, ML algorithms can learn complex relationships and patterns that might be hard to capture through traditional methods. Besides, ML techniques can handle nonlinearities and uncertainties in a more flexible manner. Furthermore, the ML algorithms have the capability to handle data that is high-dimensional and heterogeneous, enabling a more effective prediction of the melt pool in MAM processes \cite{guo2022machine, lee2019data}.

Various applications of data driven ML techniques are found in the existing literature related to melt pool analysis in MAM \cite{gaikwad2020heterogeneous, lee2019data, ogoke2021thermal, scime2019using, yuan2018machine}. Studies have employed methods like extreme gradient boosting (XGBoost) and long short-term memory (LSTM) for predicting melt pool temperature in directed energy deposition processes \cite{zhang2021prediction}. K-Nearest Neighbor (KNN) models have been used to predict porosity in samples, utilizing features extracted from real-time monitored melt pool images, demonstrating high levels of accuracy \cite{khanzadeh2018porosity}. Unsupervised ML methods, combined with computer vision, have been applied to identify keyholing porosity and balling instabilities by analyzing images from high-speed cameras during the laser powder bed fusion process \cite{scime2019using}. Artificial Neural Networks (ANN) have been found effective in correlating process parameters with melt pool features \cite{ye2023predictions}. Additionally, predictive models for 316L stainless steel have been developed using data from pyrometers and high-speed video cameras, with a Sequential Decision Analysis Neural Network (SeDANN) model showing promising results in predicting melt pool width \cite{gaikwad2020heterogeneous}.

Deep Learning (DL) based methods have also gained traction in this area recently. Convolutional Neural Networks (CNN) with transfer learning and Residual-Recurrent Convolutional Neural Networks (Res-RCNN) have been employed for real-time porosity prediction using thermal melt pool images \cite{ho2021dlam}. Deep Neural Network (DNN) architectures have been proposed for predicting melt pool geometries during in-situ processes through linear regression models \cite{milaat2021prediction}. Furthermore, deep learning methods based on CNN have been used for real-time melt pool classification, enabling quick responses to unexpected melt pool behaviors \cite{yang2019investigation}.

While ML models offer a simpler approach to application, several challenges arise when using them for melt pool prediction in Metal Additive Manufacturing (MAM). First, many ML methods, such as Neural Networks, face interpretability issues, making it difficult to understand how decisions are made \cite{gilpin2018explaining}. Second, the effectiveness of these methods in limited data scenarios is often questionable. Insufficient data can lead to overfitting in predictive models, compromising their ability to generalize well \cite{yoshida2017spectral}. Third, typical ML approaches in MAM usually provide precise predictions without accounting for uncertainty. In critical applications where inaccuracies can have severe repercussions, incorporating uncertainty into predictions is vital \cite{mondal2020investigation}. Lastly, developing an accurate ML-based prediction model requires initial training with extensive experimental data, which can be both expensive and time-intensive \cite{saunders2022additive}.

Sequential learning-based experimental designs offer a compelling solution to many of the issues commonly encountered by traditional ML algorithms in predicting melt pool geometry in MAM. The significance of these sequential experiment design strategies in accurately characterizing melt pool geometry cannot be overstated. One of the key advantages of algorithms based on sequential experiment design is their reduced reliance on large datasets for training. These models can be initially constructed with a relatively small dataset and then progressively updated as more data becomes available. This approach not only makes the models highly adaptable but also significantly decreases uncertainty. Furthermore, these models are designed to intelligently determine which subsequent data points are most beneficial for updating the model to predict the desired outcomes, requiring fewer experiments or observed data points. This feature is particularly valuable in minimizing costs, especially in resource-limited settings. In recent years, researchers are concentrating more in sequential experiment design to get a deeper knowledge of the complex systems under resource-constrained situations. Leveraging on the advantages of active learning, research in the field of material science and manufacturing is undergoing a major paradigm shift. Autonomous experimentation platforms, capable of accelerating the process of discovering new materials or finding optimum conditions for developing materials with desired property and quality, are being proposed and developed by researchers \cite{bukkapatnam2023autonomous,burger2020mobile,nikolaev2016autonomy}. One of the most popular and widely adopted sequential strategies in developing autonomous experimentation platforms is the Bayesian Optimization (BO) framework \cite{talapatra2018autonomous}. Based on Bayesian statistics and decision theory, BO uses machine learning to build a predictive surrogate model between the design parameters of a material (process parameters in this case) and its properties (melt pool geometry in this case), and then uses decision theory to suggest which design or designs (process parameters) would be most valuable to evaluate next \cite{frazier2015bayesian}. In the traditional BO framework, Gaussian Process (GP) is routinely used as the surrogate model to approximate the true underlying relationship between the input features and the target labels \cite{liu2013gaussian, perdikaris2015multi, ramakrishnan2005gaussian}. There have been several studies in AM which utilize GP regression techniques to initialize and update the surrogate models \cite{saunders2022additive, zhu2018machine, kamath2016data, meng2020process, tapia2018gaussian, lee2020optimization}. The other component of BO is the use acquisition functions for making the key decision of selecting the locations of the experiments that should be carried out sequentially to predict the underlying relationship. When making such a decision, an inevitable trade-off between exploration and exploitation exists. Among the various acquisition functions (also known as utility functions) discussed in the literature, Expected Improvement (EI) and upper confidence bound (UCB) are two widely used ones. However, these acquisition functions used in the BO framework are often considered greedy as their attempt to get to the optima are often disrupted by their over exploitative and over-explorative nature respectively \cite{bull2011convergence, chen2023hierarchical}. Despite the extensive scope of additive manufacturing, research focusing on sequential or active learning methods remains relatively limited compared to other data-driven approaches. Moreover, the majority of studies in AM that use BO as the active learning framework deal with acquisition functions (such as EI and UCB) which are biased either towards exploitation or exploration, making it difficult to reach the optimal solution with fewer iterations or evaluations \cite{deneault2021toward, mondal2020investigation, liu2022nonparametric, zhang2021bayesian, goguelin2021bayesian, xiong2019data}. Consequently, BO frameworks are yet to see an acquisition function that balances the exploration and exploitation capability of the sequential framework.

In this study, we propose a novel surprise driven sequential experiment design strategy. Based on GP and two existing surprise metrics in the literature, the sequential experiments in our proposed approach will be conducted upon observing and reacting to surprises. We claim that upon guided by surprises or observations that do not agree with the current belief or hypothesis, an ideal balance between exploration and exploitation can be achieved. In predicting the melt pool behavior, such an approach can effectively determine the true underlying relationship between the process parameters and melt pool geometry with a limited number of experiments. 

\section{Methodology}
\label{sec:sample3}

In this section, we outline the methodology behind our two proposed frameworks, which are structured into two distinct phases. Initially, in Phase I, we focus on evaluating the performance of our newly developed surprise-guided acquisition function-based sequential learning framework (SurpriseAF-BO). This is compared against the traditional EI-based BO method and six other prevalent machine learning techniques, specifically in the context of predicting melt pool geometry. Moving to Phase II, we introduce a hybrid framework (CT-SurpriseAF-BO), which enhances the original phase one framework by incorporating a CTGAN into the original framework. This integration aims to further improve the efficacy of the surprise-guided methods.

Our approach is rooted in the active learning framework commonly used in traditional BO. Therefore, we begin by discussing the BO framework itself, particularly focusing on GP and EI, which serve as the surrogate model and acquisition functions, respectively. Following this, we delve into the specifics of our proposed SurpriseAF-BO framework, which is centered around a surprise-guided acquisition function. Lastly, we elaborate on the CT-SurpriseAF-BO framework, employed in Phase II, which is designed to enhance the predictive accuracy regarding the melt pool's depth, width, and length.

\subsection{Sequential Experiment Design with Bayesian Optimization}

BO is a sequential model-based optimization technique used to find the optimal values of an objective function that may be expensive to evaluate. It consists of two components: a surrogate model and an acquisition function. 

\subsubsection{Surrogate Model}

In the context of sequential experimentation using the BO framework, it is essential to employ a surrogate model. This model is designed to initially establish an understanding of the design space and then continuously refine this understanding as it assimilates new observational data. Gaussian process (GP) is the most widely implemented surrogate model in the BO framework. The adoption of GP allows for better alignment and comparability with existing work within the BO framework, such as the Expected Improvement (EI) criterion and its variations. GP models are nonparametric, providing significant flexibility and adaptability for modeling complex response surfaces. They are particularly effective in the sequential learning framework, often serving as the go-to surrogate for black-box functions \cite{frazier2018tutorial}. A GP is defined as a stochastic process where each point $\mathbf {x} \in \mathbb{R}^{P}$ (with $P$ being the number of parameters) is linked to a random variable $f(\mathbf {x})$. Any finite set of these random variables adheres to a multivariate Gaussian distribution, which is expressed as follows:


\begin{equation}
\text{GP} \sim p(\mathbf{f}|\mathbf{X}) = \mathcal{N}(\mathbf{f}|\bm{\mu}, \mathbf{K})
\end{equation}

In the above equation, $\mathbf{f}= {f(\mathbf{x_1}), \ldots, f(\mathbf{x_n})}$ represents the set of function values, $\bm{\mu}=m(\mathbf{X})={m(\mathbf{x_1}), \ldots, m(\mathbf{x_n})}$ is the mean function, and $\mathbf{K}=k(\mathbf{x_i}, \mathbf{x_j})$ denotes the covariance function. There are various popular covariance functions used in GP models, such as the squared exponential, the Matern class, the rational quadratic, and the Ornstein-Uhlenbeck covariance functions \cite{rasmussen2006gaussian}. 

The covariance functions are key in shaping the Gaussian Process (GP) model's ability to accurately represent the complexities of the design space. In the context of using GP within a sequential learning framework for physical experiments, the function $f(\mathbf{x})$ being modeled is presumed to have specific characteristics \cite{rasmussen2006gaussian}. These include being continuous and costly to evaluate, implying that their assessments are limited due to significant time or financial constraints. The functions are treated as black-box functions, meaning they do not have a discernible special structure. They are also derivative-free, which means that we only have access to the observations of $f(\mathbf{x})$ without any information on its first or second-order derivatives. Moreover, the actual physical response $\mathbf{y}$ is typically noisy and deviates from $f(\mathbf{x})$ due to independent Gaussian noise with constant variance.

Within the GP framework, a prior distribution is assigned to the function space, which the sequential algorithm seeks to learn from, using data gathered through sequential experiments. The covariance function, $\mathbf{K}$, plays a crucial role in defining this prior, as it determines the degree of similarity between pairs of experimental data points. Essentially, the covariance function ensures that if two points, $\mathbf{x_{i}}$ and $\mathbf{x_{j}}$, are close to each other, their corresponding function evaluations, ($f(\mathbf{x_{i}}),f(\mathbf{x_{j}})$), are likely to be similar \cite{brochu2010tutorial}. For example, in the case of the squared exponential covariance function, the similarity between two function evaluations is quantified based on how close these two points are to each other as follows:

\begin{equation}
    k(\mathbf{x_i}, \mathbf{x_j}) = \sigma_s^2 \exp\left(-\frac{\|\mathbf{x_i} - \mathbf{x_j}\|_2^2}{2l^2}\right)
\end{equation}

This function encompasses two critical parameters that need to be estimated during the sequential learning process: $l$, the length scale parameter, which influences the smoothness of the functional representation of the design space, and $\sigma_s$, the variance parameter, which determines the scale of the function values. These parameters, collectively referred to as $\phi$, are the hyperparameters of the selected covariance function and are fundamental to the statistical model. The GP serves as a surrogate model for the actual design space and undergoes iterative refinement through a probabilistic Bayesian approach, consistent with the principles of sequential experimentation. Initially, a series of preliminary experiments denoted as $\mathbf{X}$, are conducted to gain an initial understanding of the material design space. Following the completion of these experiments, the GP model is trained with this data. The hyperparameters of the model, $\phi$, are then determined by maximizing the log marginal likelihood, as detailed in the subsequent equation:

\begin{align}
    \max \log (p(\mathbf{y} \mid \mathbf{X})) &= -\frac{1}{2} (\mathbf{y} - m(\mathbf{X}))^T \mathbf{K}_y^{-1} (\mathbf{y} - m(\mathbf{X})) \nonumber \\
    &\quad - \frac{1}{2} \log |\mathbf{K}_y| - \frac{n}{2} \log 2\pi
\end{align}

In equation (3), \( \mathbf{K}_y \) is the covariance matrix where, \( \mathbf{K}_y = \mathbf{K} + \sigma^2 \mathbf{I} \). \( \sigma^2 \) represents the variance of the noisy response values, and \( |\mathbf{K}_y| \) denotes the determinant of \( \mathbf{K}_y \). Predicting the experimental outcome at new test locations, \( \mathbf{X}_* \) is crucial for comparing the predicted values with the actual outcomes, \( \mathbf{y}_* \) and updating the sequential policy after each iteration. The prediction is carried out using the model's hyperparameters estimated from the previous round of experimental data. GP simplifies this prediction task, as the posterior predictive distribution of the function response at the new locations, \( \mathbf{f}_* \), is given by:

\begin{equation}
    p(\mathbf{f}_* \mid \mathbf{f}, \mathbf{X}_*, \mathbf{X}) \sim \mathcal{N}(\mathbf{f}_* \mid \bm{\mu}_*, \mathbf{\Sigma}_*)
\end{equation}

In equation (4), \( \bm{\mu}_* = \mathbf{K}_*^T \mathbf{K}_y^{-1} f \), \( \bm{\Sigma}_* = \mathbf{K}_{**} - \mathbf{K}_*^T \mathbf{K}_y^{-1} \), $\mathbf K_{*}=k(\mathbf {X},\mathbf X_{*})$ and $\mathbf K_{**}=k(\mathbf X_{*},\mathbf X_{*})$.. The posterior distribution, \( p(\mathbf{f}_* \mid \mathbf{f}, \mathbf{X}_*, \mathbf{X}) \) represents the updated understanding of the design space and can be iteratively refined with each new experiment \cite{rasmussen2006gaussian}.

\subsubsection{Acquisition Function}

In sequential learning processes, BO aims to determine the optimal values of input variables or process parameters, denoted as \( \mathbf{x}_{\text{opt}} \), in order to achieve the global maximum or minimum of the target property within the MDS. Its objective is to efficiently search for the best possible configuration that maximizes (or minimizes) the target function, which is expressed by the following equation.

\begin{equation}
    \mathbf{x}_{\text{opt}} = \underset{\mathbf{x} \in \mathbb{R}^P}{\arg\max} \, f(\mathbf{x})
\end{equation}

While the surrogate model is used to approximate the behavior of the true objective function, the purpose of the acquisition function in the BO framework is to guide the search for the optimal solution by determining the next point to evaluate. The acquisition function attempts to quantify the utility or potential of different candidate data points or experiment locations in the MDS based on the available information from the surrogate model. The acquisition function balances the exploration-exploitation trade-off. By taking into account both the surrogate model's predictions (exploitation) and the uncertainty or variability in those predictions (exploration), it tries to achieve a balance between exploration and exploitation. It considers areas in the MDS where the surrogate model is uncertain or predicts high potential and directs the optimization algorithm to explore unexplored regions that may contain better solutions. At the same time, it also exploits regions where the surrogate model predicts high performance. Proposed by Jones et al., EI is one of the most frequently used acquisition functions in the BO framework \cite{jones1998efficient}. 

EI aims to find a balance between exploitation and exploration and has been empirically demonstrated to be effective in guiding the sequential search in BO. It is expressed by the equation below.

\begin{equation}
    \text{EI}_n(\mathbf{x}) = (\mu_*(\mathbf{x}) - f(\mathbf{x}^+))\Phi\left(\frac{\Delta_n(\mathbf{x})}{\sigma_*(\mathbf{x})}\right) + \sigma_*(\mathbf{x})\phi\left(\frac{\Delta_n(\mathbf{x})}{\sigma_*(\mathbf{x})}\right)
\end{equation}

In equation (6), $(\Delta_n(\mathbf{x}) = \mu_*(\mathbf{x}) - f(\mathbf{x}^+))$ represents the potential improvement over the current best solution, $(\mathbf{x}^+)$; where \(\mu_*(\mathbf{x})\) and $(\sigma_*(\mathbf{x})$) correspond to the mean and standard deviation of the GP posterior predictive at point $\mathbf{x}$, respectively. Besides, $\Phi$ and $\phi$ represent the cumulative distribution function (CDF) and probability density function (PDF) of the standard normal distribution, respectively.

\begin{figure*}[h]
\centering
\includegraphics[width=0.8\linewidth]{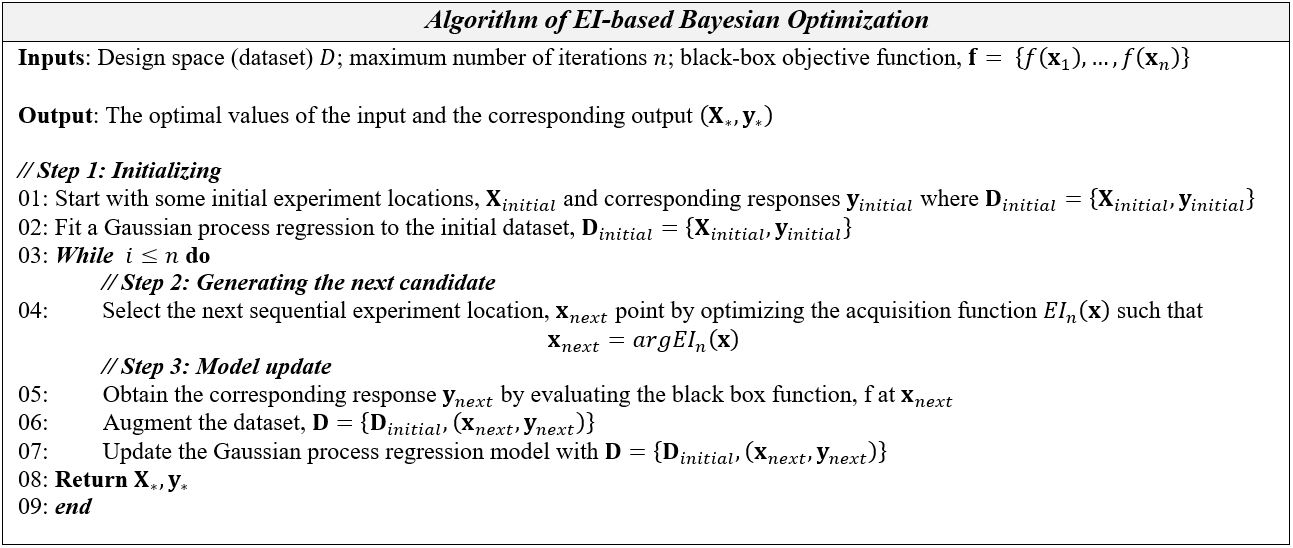}
\caption{Algorithm of EI-based Bayesian optimization}
\label{fig:myimage1}
\end{figure*}

Equation 6 is the summation of two different parts where the first part prioritizes exploiting existing knowledge and the second part emphasizes exploring new locations in the MDS. The exploitation-exploration tradeoff is balanced by optimizing this acquisition function and consequently, the next location for the sequential experiment is chosen. The pseudocode for the EI-based BO algorithm is shown in Figure~\ref{fig:myimage1}.

\begin{figure}[h]
\centering
\includegraphics[width=0.95\linewidth]{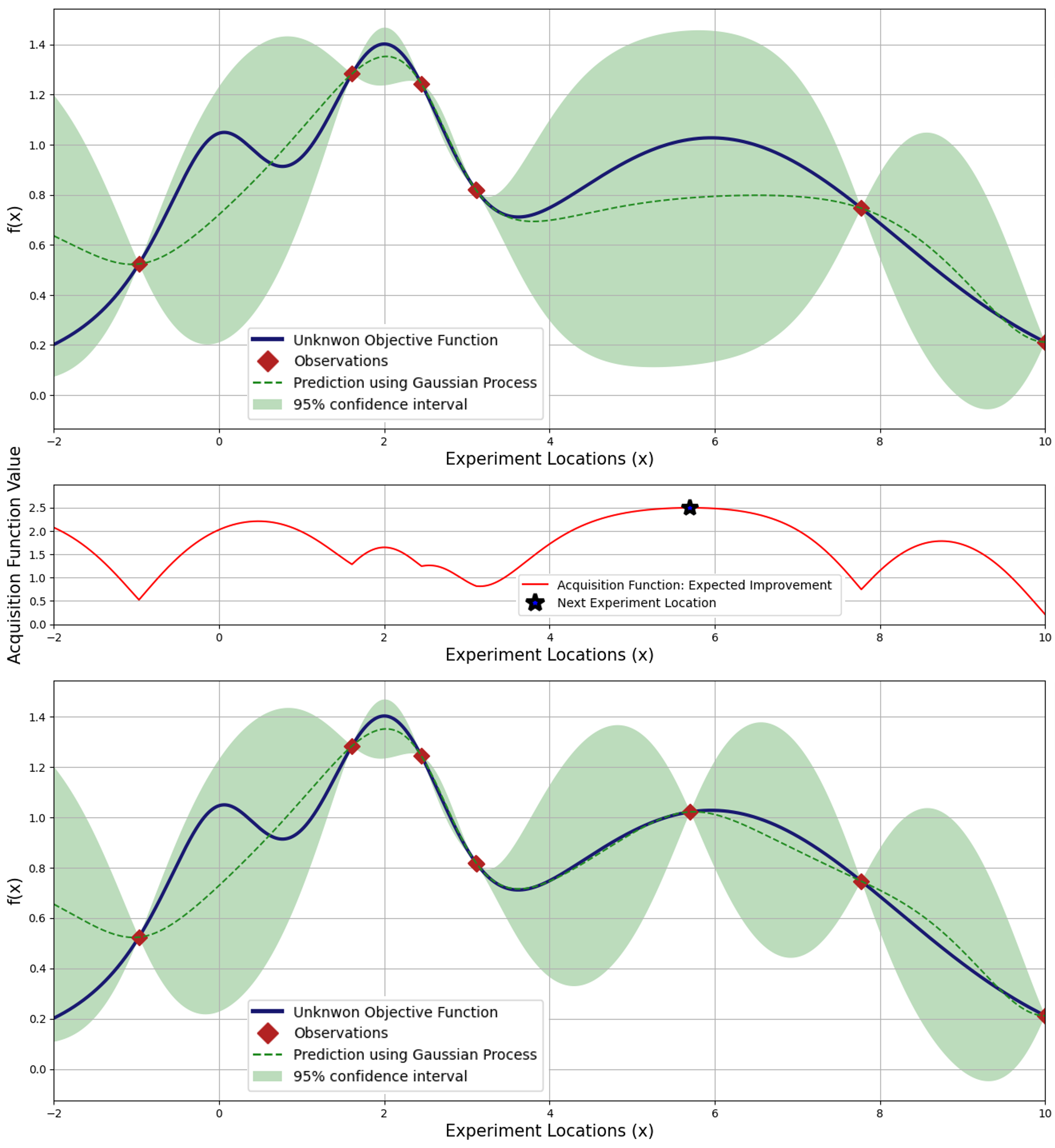}
\caption{Visualization of sequential learning process through Bayesian optimization}
\label{fig:myimage2}
\end{figure}

Figure~\ref{fig:myimage2} illustrates the process of BO. The plot at the top shows the initial approximation of the unknown function using GP with some initial samples. The plot in the middle shows the location of the next sequential sample by maximizing the value of the EI-based acquisition function. The plot at the bottom shows how the model is being updated after conducting an experiment at the selected location.

\subsection{Proposed Framework with Surprise-guided Acquisition Function (SurpiseAF-BO)}

We introduce a novel enhancement to the traditional BO method by incorporating a surprise-guided acquisition policy which allows us to refine and adapt our understanding dynamically based on the experimental outcomes.

\subsubsection{Understanding the Context of `Surprise'}

Surprise is essentially an encounter with observations that challenge or contradict our existing hypotheses or understanding of a system \cite{faraji2018balancing}. It triggers a sense of bewilderment, prompting an in-depth exploration to understand the anomaly. This pursuit often leads to a reevaluation of our current knowledge about the system or process in question, paving the way for deeper insights and understanding. In scientific contexts, the emergence of surprise naturally leads to a focused search of the area immediately surrounding the unexpected occurrence. This is done to decipher the nature and scope of the surprising responses. Engaging in this investigation of surprising observations is key to discovering new local patterns within the response surface, enhancing our comprehension of the system \cite{ahmed2021towards, jin2022autonomous}.

\subsubsection{Metrics to Quantify Surprise}

The surprise-guided acquisition function in our study is designed to accurately identify surprising observations and adapt the data collection strategy accordingly. To tackle the challenge of accurately quantifying surprise, we have adopted two well-established metrics from the realms of information science and computer science. These metrics, known as Shannon surprise and Bayesian surprise, are instrumental in quantifying the abstract concept of surprise \cite{baldi2002computational,itti2005bayesian}.

The Shannon surprise metric evaluates the unexpectedness of an observation based on our existing beliefs. It is calculated as the negative log-likelihood of an observation, where a lower probability of occurrence indicates a higher level of surprise. In contrast, the Bayesian surprise metric focuses on the extent to which a new observation alters our existing beliefs. This is measured using the Kullback-Leibler (KL) divergence, with a greater divergence signifying a more significant surprise \cite{zeng2014detecting}. Table~\ref{tab:surprise_metrics} provides detailed information about these two measures of surprise.

\newcolumntype{M}[1]{>{\centering\arraybackslash}m{#1}}

\begin{table*}[h]
\centering
\caption{A summary of Shannon and Bayesian surprise metrics used to quantify surprise}
\label{tab:surprise_metrics}
\begin{tabular}{|M{0.2\linewidth}|p{0.35\linewidth}|p{0.35\linewidth}|}
\hline
\textbf{Basis of comparison} & \textbf{Shannon surprise metric} & \textbf{Bayesian surprise metric} \\
\hline
\textbf{Idea} & Compare a new observation with the current hypothesis about the underlying system captured by a statistical model after observing \( n \) data points \( \pi_n(\bm{\theta}) \)
 &
Change in an individual's belief due to a new observation, where \( \pi_{n+1}(\bm{\theta}) \) represents the revised belief obtained after observing the new data point, \( \mathbf{D} \). \\
\hline
\textbf{Main Concept} & 
Calculates the negative log-likelihood of an observation, \( \mathbf{D} = \{\mathbf{x}, \mathbf{y}\} \) where \( \mathbf{x} \in \mathbb{R}^P \). & 
Measures the alteration in belief using the KL divergence between prior and posterior distributions. \\
\hline
\textbf{Defining Equation} & 
Expressed as:
\begin{equation}
    -\log \int_{\bm{\theta}} p(\mathbf{D} \mid \bm{\theta}) \pi_n(\bm{\theta}) \, d\bm{\theta} \quad
\end{equation}
where the probability of a new data point, given the parameterization \( \bm{\theta} \), is denoted as \( p(\mathbf{D} \mid \bm{\theta}) \). It is important to note that, \( \bm{\theta} \) and \( \pi_n(\bm{\theta}) \) are equivalent to the previously discussed \( \mathbf{f} \), and the GP posterior \( p(\mathbf{f}_* \mid \mathbf{f},\mathbf{X}_*,\mathbf{X}) \), respectively described in Equation 4.
 &
Expressed as:
\begin{equation}
    KL(\pi_n(\bm{\theta}) \| \pi_{n+1}(\bm{\theta})) \quad
\end{equation}
where KL divergence measures the disparity between the updated model \( \pi_{n+1}(\bm{\theta}) \) and the previous model \( \pi_n(\bm{\theta}) \). Also, \( \pi_{n+1}(\bm{\theta}) \) is calculated as follows using Bayes' rule.
\begin{equation}
    \pi_{n+1}(\bm{\theta}) = \frac{p(\mathbf{D} \mid \bm{\theta}) \pi_n(\bm{\theta})}{\int_{\bm{\theta}} p(\mathbf{D} \mid \bm{\theta}) \pi_n(\bm{\theta}) \, d\bm{\theta}}
\end{equation}. \\
\hline
\textbf{Explanation} & 
When the observation has a low probability, it indicates a higher surprise. Not all low probability events are surprising, though. & 
After observing a new point, the updated model is obtained and compared to the previous model using KL divergence. A significant KL divergence is a marker of a substantial surprise. \\
\hline
\end{tabular}
\end{table*}

\subsubsection{Working Flow of SurpiseAF-BO}

\textbf{Stage 1: Identification of Surprise:} Once the surprise levels are quantified using Shannon and Bayesian measures, the next step is to effectively identify and react to these surprising observations. This requires setting a threshold to determine what constitutes a surprise. Our approach involves using a credible interval, denoted as $\bm{\mu}_{*}\pm k_{\text{Shannon}} \mathbf{\Sigma}_{*}$, for each new test response ($\mathbf{y}_{*}$). For example, in a scenario where the credible interval is 95\%, $k_{\text{Shannon}}$ would be set to 1.96 for a normal distribution. An observation is considered a surprise if its level of surprise surpasses the confidence bound, indicating a discrepancy between the new test result and the existing model predictions. In a similar vein, for Bayesian surprise, we employ a corresponding factor, $k_\text{Bayesian}$, which serves a role analogous to $k_\text{Shannon}$.

\textbf{Stage 2: Validation of Surprise – An Act of Exploitation:} When a surprise observation arises, it is crucial to determine the appropriate response. Such an observation might stem from a corrupted or noisy data point, making it necessary to verify its legitimacy before updating the model based on this potentially misleading information. To address this, we implement a confirmation step. This step involves collecting a new observation near the location of the initial surprise to see if it also registers as a surprise. If this subsequent observation is similarly flagged as a surprise, it lends credibility to the original surprise observation, and both data points are then considered valid and retained. On the other hand, if the new observation does not result in a surprise, the initial observation may be deemed an anomaly or error and consequently discarded. This validation process, conducted near the original surprise, represents a focused effort at exploitation. It is designed to refine the decision-making process by ensuring that only genuine surprise observations, indicative of true deviations or novel insights, are used to update the model.

\textbf{Stage 3: Iterative Refinement and Exploration:} Following the confirmation of surprise in Stage 2, additional observations are collected from the same local neighborhood to update the model, continuing until a new data collection no longer yields a surprise observation. This indicates that the model has achieved consistency with the information provided by the data for that specific local area. Upon completing this local exploitation phase, the model shifts its focus to exploration. This involves selecting the next observation or experiment location, denoted as $\mathbf{x}_{\text{next}}$. The selection is guided by the maximin policy, which is designed to maximize the minimum expected improvement \cite{joseph2015maximum}. This policy is articulated in equation (10), where the objective is to identify a candidate location that maximizes the minimum distance between the new design location and the locations of previously used designs. This approach ensures a strategic balance between refining the model based on new, surprising data and exploring new areas to enhance the model's overall understanding and performance.

\begin{equation}
    \mathbf{x}_{\text{next}} = \underset{\mathbf{x} \in \mathbf{S}}{\arg\max} \, G(\mathbf{x}) \quad (9)
\end{equation}

In the above equation, $\mathbf{S}$ represents the experiment budget (available candidate locations) for the sequential experiments, \( G(\mathbf{x}) = \min_{\mathbf{e} \in \mathbf{E}} \|\mathbf{x} - \mathbf{e}\|_2 \) calculates the minimum distance between a given location, $\mathbf{x}$, in the candidate set from any of the already used locations denoted by $\mathbf{E}$.

\begin{figure*}[h]
\centering
\includegraphics[width=0.85\linewidth]{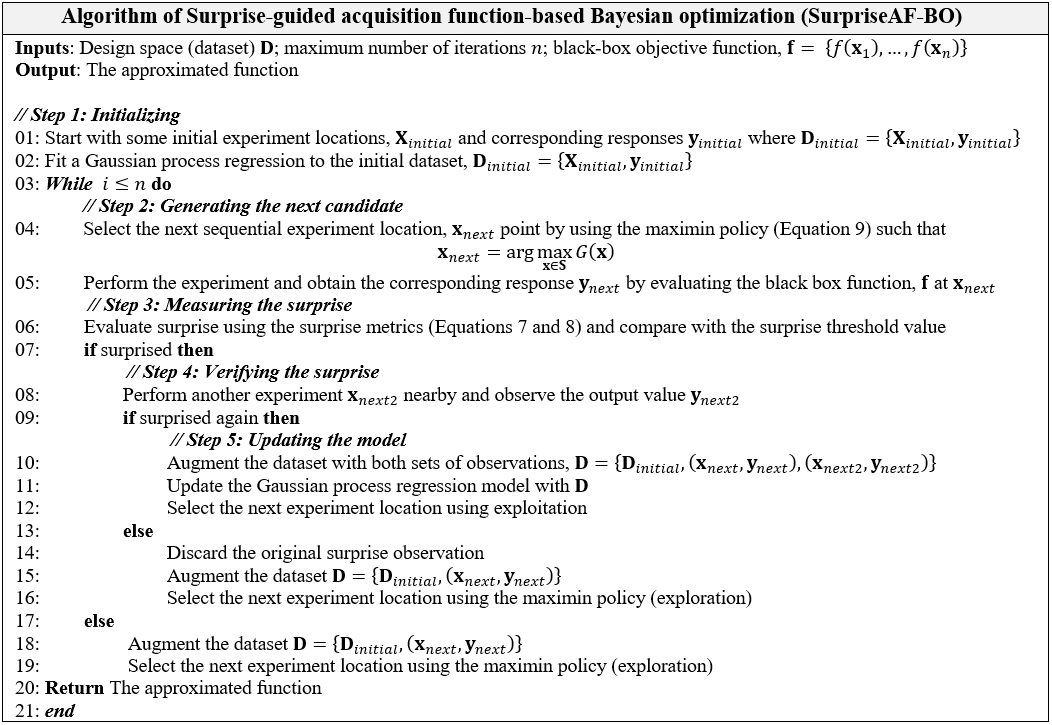}
\caption{Algorithm of Surprise-guided acquisition function-based Bayesian optimization (SurpriseAF-BO)}
\label{fig:myimage3}
\end{figure*}

Now, the degree of surprise associated with this new observation will be measured once again according to Stage 1. If no surprise is encountered, the exploration phase will continue using the same maximin policy until a surprise is encountered. Once a surprise is observed, the exploitation phase will commence as mentioned in Stage 2. This iterative process will continue until the experimentation budget is depleted. The algorithm of the sequential learning framework with surprise guided acquisition function is shown in Figure~\ref{fig:myimage3} where its working mechanism can be illustrated in five steps. The algorithm is initialized by fitting a GP with \( \mathbf{D}_{\text{initial}} \) samples or experiments. Steps 2 and 5 correspond to Stage 3 discussed above. Steps 3 and 4 resemble Stages 1 and 2, respectively.

\subsection{Conditional Generative Adversarial Network (CTGAN)}

In data-scarce and resource-constrained settings, the significance of synthetic data becomes paramount. It serves as a foundation for training, testing, and validating machine learning models, especially when real-world data acquisition is challenging or not feasible. Synthetic data augments existing datasets, enhancing their volume and diversity, which is vital for preventing overfitting and bolstering model generalization, especially in deep learning. Beyond its role in model development, synthetic data offers privacy advantages by emulating real data's statistical features without revealing sensitive information, making it a safe choice for research sharing. Economically, it provides a cost-effective solution by eliminating the hefty expenses related to real-world data collection, like sensors and manual annotations. Additionally, it facilitates the simulation of specific or rare scenarios, invaluable in areas like disaster response, and addresses imbalances in data by supplementing underrepresented classes, ensuring holistic model training.

Generative adversarial networks (GANs) have emerged as one of the most effective tools for generating synthetic data \cite{fang2022dp}. The GAN framework involves two networks, the generator ($Gen$), and the discriminator ($Dis$), which are trained in tandem. The generator aims to produce data indistinguishable from real data, while the discriminator's task is to distinguish between genuine and generated data. This adversarial process results in the generator producing increasingly realistic data over time. The following equation shows the objective function that needs to be minimized during the learning process of GAN.

\begin{align}
    \min_{Gen} \max_{Dis} V(Dis, Gen) &= \mathbb{E}_{x \sim p_{\text{data}}(x)}[\log Dis(x)] \nonumber \\
    &\quad + \mathbb{E}_{z \sim p_{z}(z)}[\log (1 - Dis(Gen(z)))] \quad
\end{align}

In the above equation, \( p_{\text{data}}(x) \) represents real data, while \( p_{z}(z) \) signifies the synthetic data. The discriminator, \( Dis \), takes real data input \( x \) and evaluates its authenticity, producing a probability score, \( Dis(x) \). The generator, \( Gen \), uses a random noise vector \( z \) to generate synthetic data depicted by \( Gen(z) \). \( Dis \)'s objective is to correctly classify real data, aiming for \( Dis(x) \) to approach 1, while classifying the synthetic data, \( Dis(Gen(z)) \), as 0. Conversely, \( Gen \)'s aim is to create data so convincing that \( Dis(Gen(z)) \) approaches 1, effectively tricking \( Dis \). Hence, the equation's goal is dual, since \( Dis \) seeks to maximize its accuracy, while \( Gen \) endeavors to minimize \( Dis \)'s ability to differentiate.

GANs have revolutionized artificial data generation in various sectors \cite{habibi2023imbalanced, kas2022coarse}, including healthcare \cite{armanious2020medgan}, agriculture \cite{kerdegari2019smart}, and image processing \cite{malakshan2023joint}. However, their application in synthesizing tabular data encounters specific challenges. Originally designed for continuous data like images, GANs often struggle with tabular data, which typically comprises both continuous and categorical variables \cite{li2021improving}. A significant issue in this context is mode collapse, where GANs fail to capture the full spectrum of the real data distribution, a critical aspect in tabular data \cite{xu2019modeling}. Training GANs for tabular data can be complex, with common problems like oscillations and non-convergence, particularly due to the intricate inter-feature relationships in such data \cite{saxena2021generative}. The high-dimensional and sparse nature of tabular data also poses a challenge for traditional GAN architectures. CTGAN addresses these issues effectively. It integrates embedding layers for categorical variables and employs the WGAN-GP framework to combat mode collapse. Its architecture, designed with residual connections and normalization techniques, ensures stable training. CTGAN's support for conditional generation is a significant advancement, allowing for the creation of synthetic data under specific conditions \cite{parimala2019quality}.

CTGAN involves the use of a GAN-based method to model tabular data distribution and generate synthetic rows based on that distribution. Mode-specific normalization is incorporated into the approach to address the issues associated with non-Gaussian and multimodal distributions. Using a conditional generator, the CTGAN utilizes a training-by-sampling strategy. Several new techniques are integrated into the model's architecture in order to ensure high-quality results \cite{xu2019modeling}. Numerous studies have validated CTGAN's superiority and demonstrated its impressive performance to efficiently handle tabular data which might include both categorical and continuous variables \cite{habibi2023imbalanced, srinivas2021hardware}. While retaining the foundational learning approach of GAN, CTGAN enhances the data generation process by incorporating a feature, $y$, representing a specific condition into both G and D. This allows CTGAN to influence the generation more directly and effectively learn both the characteristics and distribution of the data. The equation representing the objective function in this case is given in the following equation.

\begin{align}
    \min_{Gen} \max_{Dis} V(Dis, Gen) &= \mathbb{E}_{x \sim p_{\text{data}}(x)}[\log Dis(x \mid y)] \nonumber \\
    &\quad + \mathbb{E}_{z \sim p_{z}(z)}[\log (1 - Dis(Gen(z \mid y)))]
\end{align}

\begin{figure*}[htbp]
\centering
\includegraphics[width=0.95\linewidth]{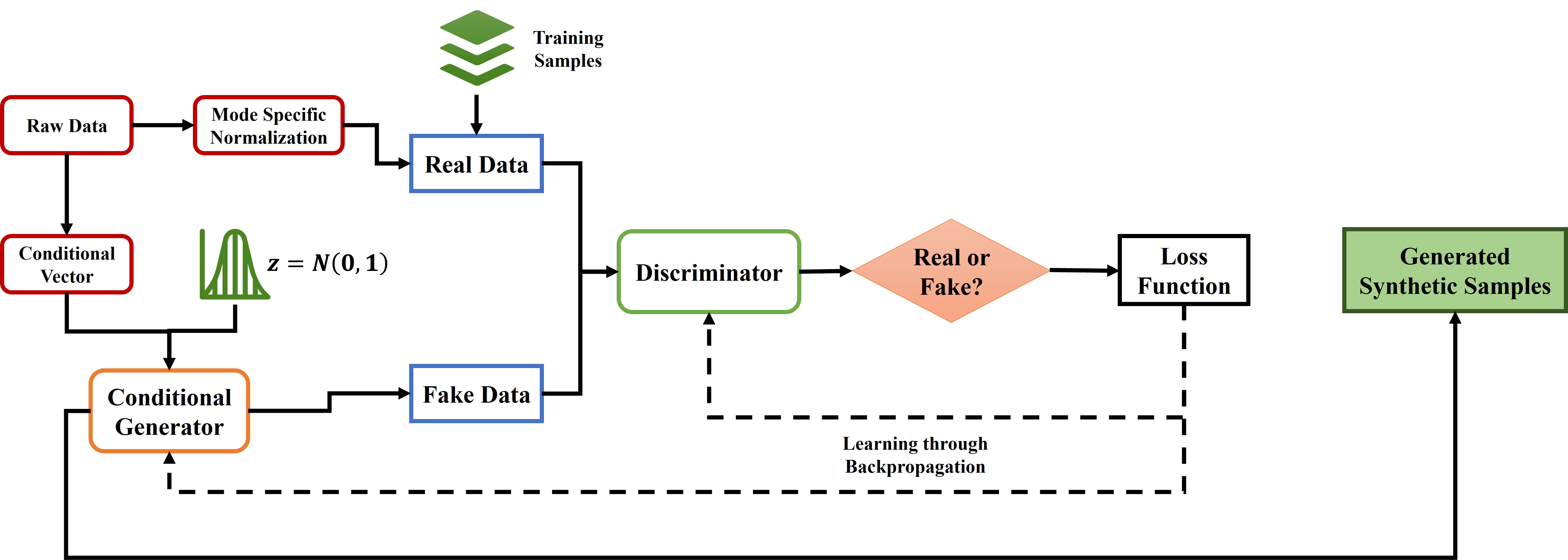}
\caption{Architecture of the CTGAN model}
\label{fig:myimage4}
\end{figure*}

Figure~\ref{fig:myimage4} illustrates the process of artificial sample generation using CTGAN.

\section{Description of the Melt Pool Dataset}
\label{sec:sample4}

This study utilizes data sourced from the MeltpoolNet, a repository established by Akbari et al. \cite{akbari2022meltpoolnet}. MeltpoolNet is a comprehensive database that includes information on melt pool geometry (depth, width, and length), defects, process parameters, and material properties, primarily gathered from published research in manufacturing and materials science journals. The data in this study specifically comes from experimental findings reported in these journals, focusing on these key aspects.

The original dataset from MeltpoolNet comprises nearly 2200 data points, encompassing details on three aspects of melt pool geometry, four types of defects, and a variety of process parameters and material properties. These data points were collected from experiments conducted using two different MAM processes: Powder Bed Fusion (PBF) and Directed Energy Deposition (DED). The dataset includes several process parameters and material properties as input features. The process parameters feature laser power, scanning velocity, beam diameter, layer thickness, and hatch spacing, while material properties include density, thermal conductivity, melting temperature, specific heat, and chemical composition. For this study, which aims to predict melt pool geometry, we selected baseline features from both process parameters and material properties relevant to L-PBF process, considering the presence of missing values. The primary target variables or output variables identified were the depth, width, and length of the melt pool. Table~\ref{tab:input_features_targets} presents a detailed list of these input features and target output variables.

\begin{table}[h]
\centering
\caption{List of six input features and three targets used in predicting melt pool geometry}
\label{tab:input_features_targets}
\begin{tabular}{|m{0.4\linewidth}|m{0.35\linewidth}|}
\hline
\multicolumn{1}{|c|}{\textbf{Input Features}} & \multicolumn{1}{c|}{\textbf{Targets}} \\ \hline
Laser power & Depth of melt pool \\ \cline{1-2}
Scanning velocity & Width of melt pool \\ \cline{1-2}
Laser beam diameter & Length of melt pool \\ \cline{1-2}
Density & \multirow{3}{*}{} \\ \cline{1-1}
Melting temperature & \\ \cline{1-1}
Thermal conductivity & \\ \hline
\end{tabular}
\end{table}

\subsection{Data Preparation Steps}
\subsubsection{Removing Missing and Duplicate Values}

In the initial dataset for the L-PBF process, there were 1294, 1086, and 293 data points for the melt pool depth, width, and length, respectively. However, this dataset contained some missing and duplicate values within these target categories. After the removal of these missing and duplicate entries, the refined dataset comprised 1115, 850, and 257 data points for the depth, width, and length of the melt pool, respectively.

\subsubsection{Normalizing the Data}

Given the diverse range of input parameters in our dataset, we employed a normalization technique to standardize the data. This process is described by the following equation:
\begin{equation}
\label{eq:norm}
x_{\text{Normalized}} = \frac{x - \bar{x}}{s}
\end{equation}
Here, $\bar{x}$ represents the sample mean, and $s$ denotes the sample standard deviation of the input parameters.

\subsubsection{Splitting the Dataset}

The dataset was divided into two parts for the purposes of model training and testing. We adopted a random splitting strategy, allocating 75\% of the data for training and the remaining 25\% for testing the models' performance. Table~\ref{tab:melt_pool_geometry} in the study details the distribution of data points in the training and testing sets for predicting the melt pool depth, width, and length.


\begin{table}[h]
\centering
\caption{Number of samples in the training and testing set for melt pool geometry prediction}
\label{tab:melt_pool_geometry}
\begin{tabular}{|c|c|c|}
\hline
\textbf{Melt pool geometry} & \textbf{Training Set} & \textbf{Testing Set} \\ \hline
Depth & 836 & 279 \\ \hline
Width & 637 & 213 \\ \hline
Length & 192 & 65 \\ \hline
\end{tabular}
\end{table}

\subsubsection{Cross Validation}

Cross-validation is a crucial technique in machine learning for assessing model performance and fine-tuning hyperparameters. It enhances the robustness and representativeness of performance evaluation by testing the model on various data subsets, thereby mitigating overfitting and bolstering generalizability. In our study, we implemented 5-fold cross-validation on the normalized data, post its division into training and testing sets.

In 5-fold cross-validation, the dataset is partitioned into five equal segments, or `folds'. For each iteration, one fold is reserved as the validation set, and the remaining four folds are amalgamated to create the training set. This cycle is repeated five times, with each fold serving as the validation set once. The model's accuracy is determined by averaging the results from these five iterations.

Furthermore, we conducted a grid search to identify the most effective hyperparameters for the ML models used in this research. By iterating the training and evaluation process across various hyperparameter combinations and assessing their average performance, 5-fold cross-validation aids in pinpointing the hyperparameters that yield the best generalization and performance across multiple validation sets. This approach ensures a more reliable and unbiased evaluation of the model's performance, facilitating optimal hyperparameter selection. Table~\ref{tab:ml_models} in the study outlines the hyperparameters selected for the different ML models used to predict melt pool geometry.

\begin{table*}[h]
\centering
\caption{List of studied and best hyperparameters for the six ML models}
\label{tab:ml_models}
\begin{tabular}{|l|l|l|l|}
\hline
\textbf{ML Model} & \textbf{Hyperparameters} & \textbf{Best Value} & \textbf{Range Studied} \\
\hline
\multirow{4}{*}{Random Forest} & n\_estimators & 200 & [20, 200] \\
\cline{2-4}
                               & max\_depth & 20 & [2,20] \\
\cline{2-4}
                               & min\_samples\_split & 2 & [2,10] \\
\cline{2-4}
                               & min\_samples\_leaf & 1 & [1,4] \\
\hline
\multirow{3}{*}{Neural Network} & hidden\_layer\_sizes & (100, 100) & (100,), (50, 50), (100, 50), \\
\cline{2-4}
                               & activation & 'relu' & 'logistic', 'tanh', 'relu' \\
\cline{2-4}
                               & solver & 'adam' & 'adam', 'sgd' \\
\hline
\multirow{3}{*}{Support Vector Regression} & kernel & 'rbf' & 'linear', 'poly', 'rbf', 'sigmoid' \\
\cline{2-4}
                               & C & 200 & [10, 200] \\
\cline{2-4}
                               & gamma & 'scale' & 'scale', 'auto' \\
\hline
\multirow{4}{*}{Gradient Boosting} & n\_estimators & 200 & [20, 200] \\
\cline{2-4}
                               & max\_depth & 20 & [2,20] \\
\cline{2-4}
                               & min\_samples\_split & 2 & [2,10] \\
\cline{2-4}
                               & min\_samples\_leaf & 1 & [1,4] \\
\hline
\multirow{2}{*}{Ridge Regression} & alpha & 0.01 & [0.01, 50] \\
\cline{2-4}
                               & solver & 'saga' & 'auto', 'svd', 'cholesky', 'saga' \\
\hline
\multirow{3}{*}{Lasso Regression} & alpha & 0.01 & [0.01, 100] \\
\cline{2-4}
                               & max\_iter & 1000 & [1000, 5000] \\
\cline{2-4}
                               & tol & 0.001 & [0.001, 0.00001] \\
\hline                               
\end{tabular}
\end{table*}

\section{Results and Discussion}
\label{sec:sample5}

\begin{figure*}[h]
\centering
\includegraphics[width=0.95\linewidth]{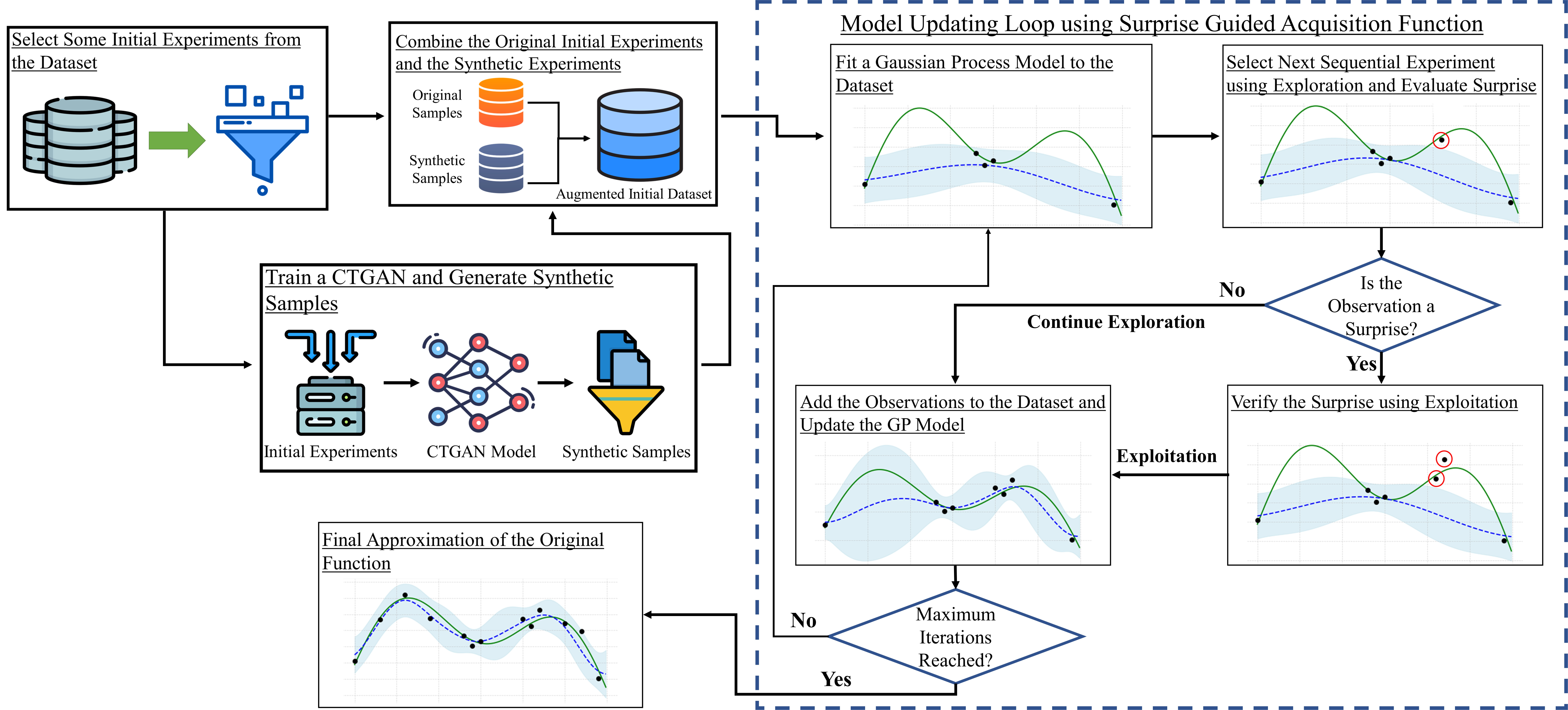}
\caption{Flow diagram of the proposed CT-SurpiseAF-BO approach}
\label{fig:myimage5}
\end{figure*}

In this section, we present the results of melt pool geometry (depth, width, and length) prediction which are divided into two phases. Phase I involves comparing the performance of SurpiseAF-BO with EI-based BO and six widely used machine learning techniques. In Phase II, we attempt to enhance SurpiseAF-BO algorithm by proposing the augmented CT-SurpiseAF-BO framework. The following sections elaborate on the results and discussions of these two phases. Figure~\ref{fig:myimage5} shows the entire flow diagram for Phase II which is for the CT-SurpiseAF-BO framework. The flow diagram for Phase I with the SurpiseAF-BO approach would be the same as Figure~\ref{fig:myimage5} without the inclusion of CTGAN and synthetic samples.

\subsection{Phase I: Comparison of SurpiseAF-BO with EI-based BO and ML Models}

In this phase, we have evaluated the performance of our proposed surprise-driven sequential learning model with the widely used EI-based BO (a sequential learning approach) and six popular ML (static approaches) techniques. As the performance metric we have incorporated root mean square error (RMSE) which is expressed by the following equation.

\begin{equation}
    RMSE = \sqrt{\frac{1}{N} \sum_{i=1}^{N} (\mathbf{y}_i - \hat{\mathbf{y}}_i)^2} \quad
\end{equation}

In equation (14), the size of the dataset is represented by $N$, $\hat{\mathbf{y}}_i$ is used to denote the predicted values after training the model, while $\mathbf{y}_i$ indicates the true values of the targets. A lower RMSE value is desired, which indicates that the predicted values are closer to the actual values, resulting in better predictive performance.

The Phase I comparison is structured in two distinct scenarios. In the first scenario, which we refer to as Scenario I, the ML algorithms are provided with double the number of data points in their training set compared to the sequential learning approaches. This setup is designed to assess the performance of the ML algorithms with a larger training dataset. Next, we conduct a second comparison, termed Scenario II or the reduced data scenario. Here, both the sequential learning models and the ML algorithms are allocated an equal number of data points in their training sets. The purpose of this scenario is to evaluate the effectiveness of the sequential learning models in environments where resources and data are limited.  

It is important to note that in our study, the training points for the ML models are randomly chosen from the available training dataset. In contrast, for the sequential learning approaches, the selection of data points is more strategic. The data points are intelligently chosen by their respective acquisition functions, which are designed to optimize resource efficiency. This distinction in data point selection methods plays a crucial role in the effectiveness and efficiency of the sequential approaches compared to the traditional ML models. Also, to ensure fairness and eliminate any potential bias in the results, all models (both sequential and traditional/static ML) are tested on the same data points each time. To accurately capture and quantify the inherent uncertainty in the RMSE values, we repeat this comparative process across 20 iterations and calculate the mean and median RMSE from these iterations. This approach provides a more comprehensive and reliable assessment of the model's performance. 

\subsubsection{Predicting Depth of Melt Pool}

\begin{figure*}[h]
\centering
\includegraphics[width=0.95\linewidth]{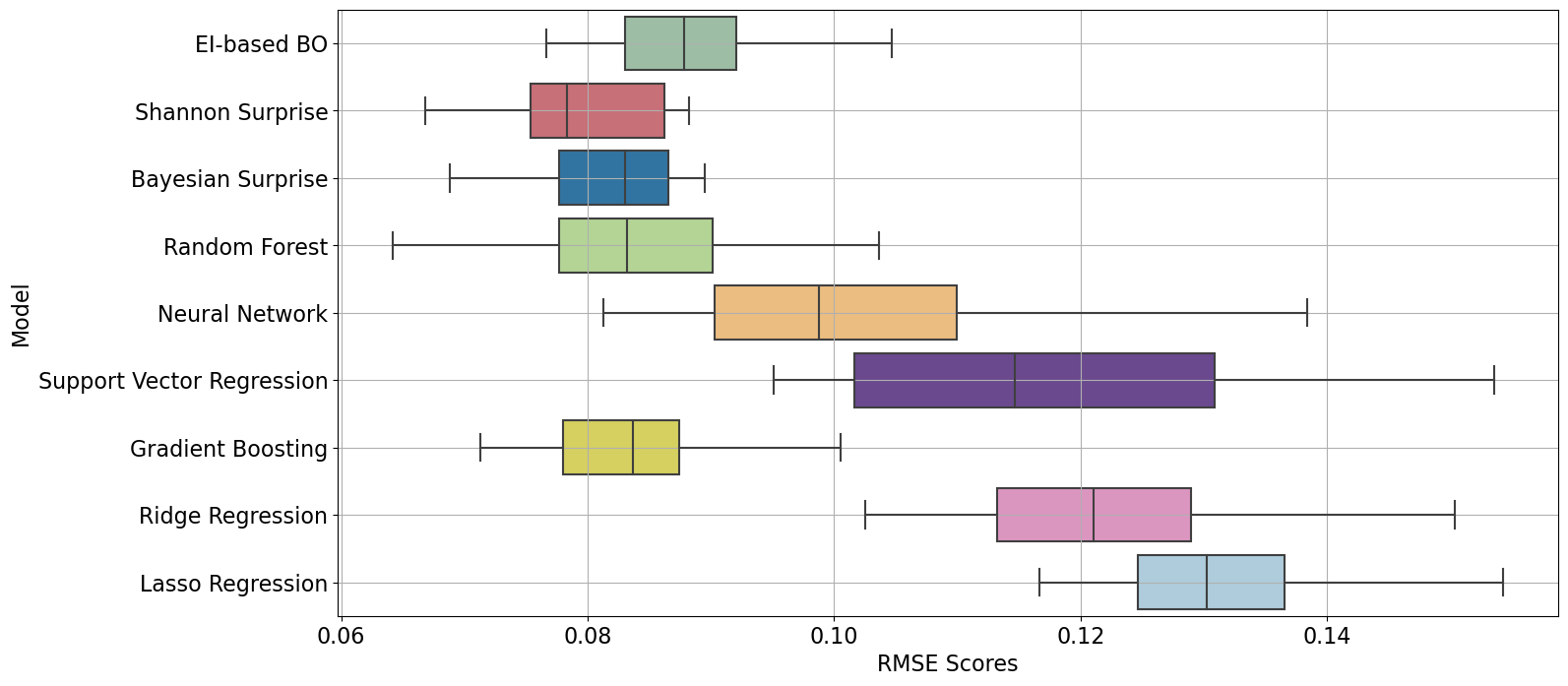}
\caption{Boxplot of RMSE of all the models in predicting the depth of melt pool (Scenario II)}
\label{fig:myimage6}
\end{figure*}

While predicting the melt pool depth, we create a set of 50 initial experiments for the two SurpiseAF-BO approaches (Shannon and Bayesian surprise) and the EI-based BO framework. For the sequential experiments, we limit the experiment budget to 175 samples or datapoints. This means, the sequential learning models are trained on a total of 225 datapoints. In contrast, the ML models are trained under two different scenarios. In the first scenario, the ML models are trained with a dataset of 450 data points, which is double the size of the dataset used for the sequential models. In the second scenario, to align with the data constraints of the sequential models, the training set for the ML models is reduced to 225 data points, matching the size of the sequential models' training set. It's important to note that the test set remains constant and identical across all models in both scenarios. The results of these experiments, including the mean and median Root Mean Square Error (RMSE) scores obtained from 20 iterations for both Scenario I and Scenario II, are detailed in Table~\ref{tab:rmse_scores}. Additionally, Figure~\ref{fig:myimage6} provides boxplots that depict the distribution of RMSE scores for all models in Scenario II, offering a visual representation of the models' performance across these scenarios.

\begin{table*}[h]
\centering
\caption{Mean and median of the RMSE scores for all the models in predicting the depth}
\label{tab:rmse_scores}
\begin{tabular}{|m{4cm}|c|c|c|c|}
\hline
\multicolumn{1}{|c|}{} & \multicolumn{2}{c|}{\textbf{Scenario I}} & \multicolumn{2}{c|}{\textbf{Scenario II}} \\ \cline{2-5} 
\multicolumn{1}{|c|}{\textbf{Model}} & \textbf{Mean} & \textbf{Median} & \textbf{Mean} & \textbf{Median} \\ \hline
EI-based BO & 0.089037 & 0.087852 & 0.089037 & 0.087852 \\ \hline
Shannon Surprise & 0.078665 & 0.078322 & \textbf{0.078665} & \textbf{0.078322} \\ \hline
Bayesian Surprise & 0.083439 & 0.083032 & 0.083439 & 0.083032 \\ \hline
Random Forest & 0.071661 & 0.068327 & 0.083597 & 0.083152 \\ \hline
Neural Network & 0.08846 & 0.0875 & 0.101274 & 0.098743 \\ \hline
Support Vector Regression & 0.096769 & 0.091495 & 0.119723 & 0.114688 \\ \hline
Gradient Boosting & 0.065087 & 0.064612 & 0.083475 & 0.083693 \\ \hline
Ridge Regression & 0.119113 & 0.117784 & 0.122413 & 0.121014 \\ \hline
Lasso Regression & 0.132807 & 0.131058 & 0.132634 & 0.130216 \\ \hline
\end{tabular}
\end{table*}

The data presented in Table~\ref{tab:rmse_scores} and Figure~\ref{fig:myimage6} provide insights into the performance of the algorithms under evaluation. In Scenario I, where the ML algorithms were trained with twice the number of data points compared to the sequential models, our proposed methods still demonstrated superior performance over most ML approaches, with the exception of Random Forest and Gradient Boosting. In Scenario II, which represents a resource-constrained environment, the Shannon surprise-guided approach excelled, outperforming all other methods, including both sequential and ML models. The Bayesian surprise-guided strategy also performed well, surpassing the ML models in terms of mean and median RMSE values, though it did not quite reach the effectiveness of the Shannon surprise-based approach. These results underscore the robustness and efficiency of our proposed sequential learning strategies, particularly in scenarios with limited data availability.

\subsubsection{Predicting Width of Melt Pool}

\begin{figure*}[h]
\centering
\includegraphics[width=0.95\linewidth]{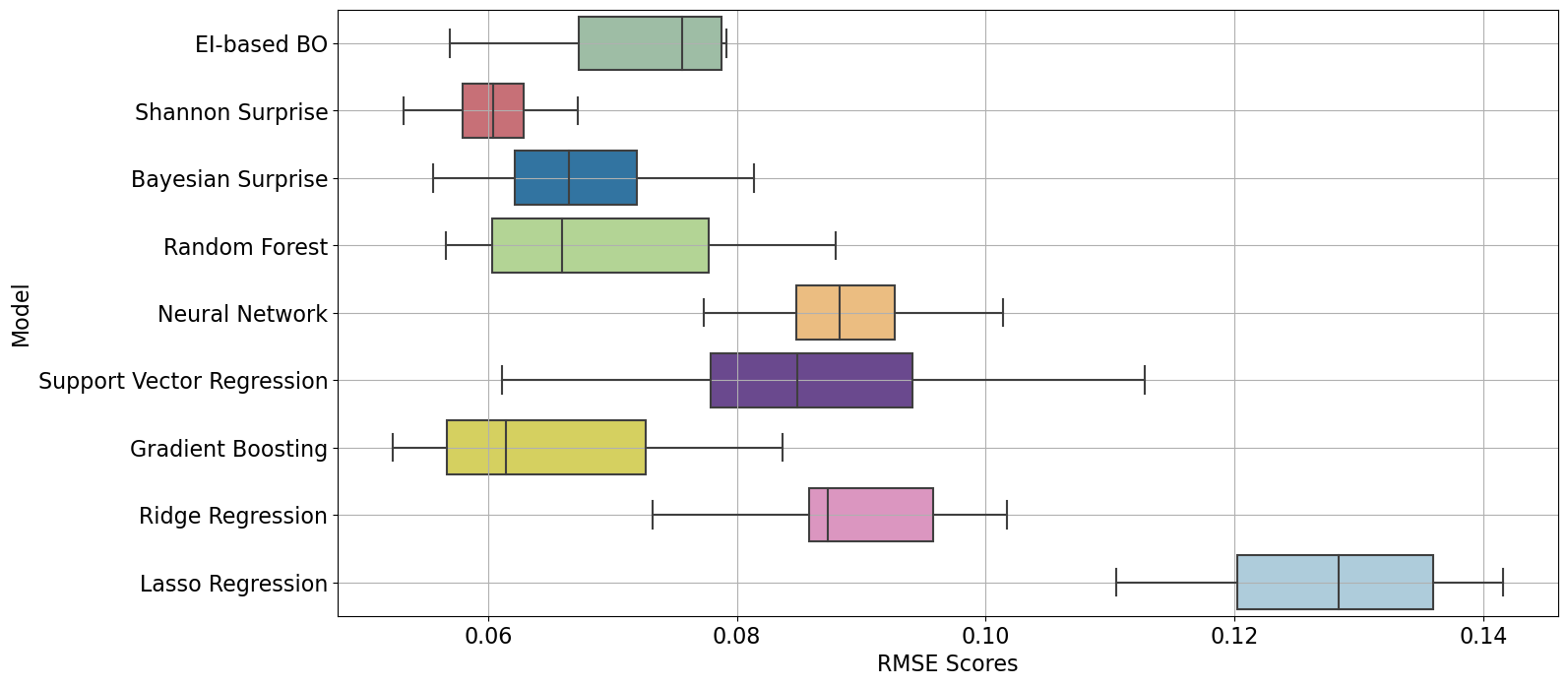}
\caption{Boxplot of RMSE of all the models in predicting the width of melt pool (Scenario II)}
\label{fig:myimage7}
\end{figure*}

In our study on predicting the width of the melt pool, we initially designated 40 data points for the initial experiments across the three sequential learning approaches. The total number of data points allocated for these sequential experiments was 125, leading to an overall count of 165 data points for training the sequential models. For the ML models, the approach varied across two scenarios. In Scenario I, the training set for the ML models was more than twice the size of that for the sequential models, totaling 350 data points. Conversely, in Scenario II, we ensured uniformity in the training set size across all models, including both the three sequential and six ML models, with each receiving 165 data points. The outcomes of these scenarios are summarized in Table~\ref{tab:rmse_scores_width}, which shows the mean and median RMSE scores from 20 iterations for both scenarios. Furthermore, Figure~\ref{fig:myimage7} visually represents these findings, displaying boxplots that depict the distribution of RMSE scores for all models in each scenario, providing a clear comparison of their performance in predicting melt pool width under different training set conditions.

\begin{table*}[h]
\centering
\caption{Mean and median of the RMSE scores for all the models in predicting the width}
\label{tab:rmse_scores_width}
\begin{tabular}{|l|c|c|c|c|}
\hline
\multicolumn{1}{|c|}{} & \multicolumn{2}{c|}{\textbf{Scenario I}} & \multicolumn{2}{c|}{\textbf{Scenario II}} \\ \cline{2-5}
\multicolumn{1}{|c|}{\textbf{Model}} & \textbf{Mean} & \textbf{Median} & \textbf{Mean} & \textbf{Median} \\ \hline
EI-based BO & 0.076133 & 0.075569 & 0.076133 & 0.075569 \\ \hline
Shannon Surprise & 0.05977 & 0.06043 & \textbf{0.05977} & \textbf{0.06043} \\ \hline
Bayesian Surprise & 0.067024 & 0.066502 & 0.067024 & 0.066502 \\ \hline
Random Forest & 0.052722 & 0.052066 & 0.068714 & 0.065909 \\ \hline
Neural Network & 0.083952 & 0.083224 & 0.088999 & 0.088242 \\ \hline
Support Vector Regression & 0.076497 & 0.075121 & 0.085319 & 0.084877 \\ \hline
Gradient Boosting & 0.052441 & 0.050456 & 0.065247 & 0.061462 \\ \hline
Ridge Regression & 0.087062 & 0.086006 & 0.089937 & 0.087273 \\ \hline
Lasso Regression & 0.128543 & 0.128791 & 0.128275 & 0.128401 \\ \hline
\end{tabular}
\end{table*}

The results from Scenario I in Table~\ref{tab:rmse_scores_width} indicate that the sequential learning methods (SurpriseAF-BO and EI-based BO) generally outperformed most of the ML techniques, with the exceptions being Random Forest and Gradient Boosting. It is important to highlight that during this phase, the sequential learning models were trained on a dataset half the size of that used for the ML models. In Scenario II, the Shannon surprise-guided approach surpassed all other methods, including Random Forest and Gradient Boosting. However, these two ML models still managed to achieve slightly better performance than the Bayesian surprise-guided strategy and the EI-based BO approach, as depicted in Figure 7. Notably, while the median RMSE score for the Shannon surprise-based method and Gradient Boosting were quite similar, the Shannon surprise-based approach exhibited significantly less dispersion in its results, indicating a more consistent performance, as can be seen in Figure 7. This suggests that the Shannon surprise-guided approach not only competes closely with top-performing ML models but also offers a more reliable and stable prediction capability in this context.

\subsubsection{Predicting Length of Melt Pool}

\begin{figure*}[h]
\centering
\includegraphics[width=0.95\linewidth]{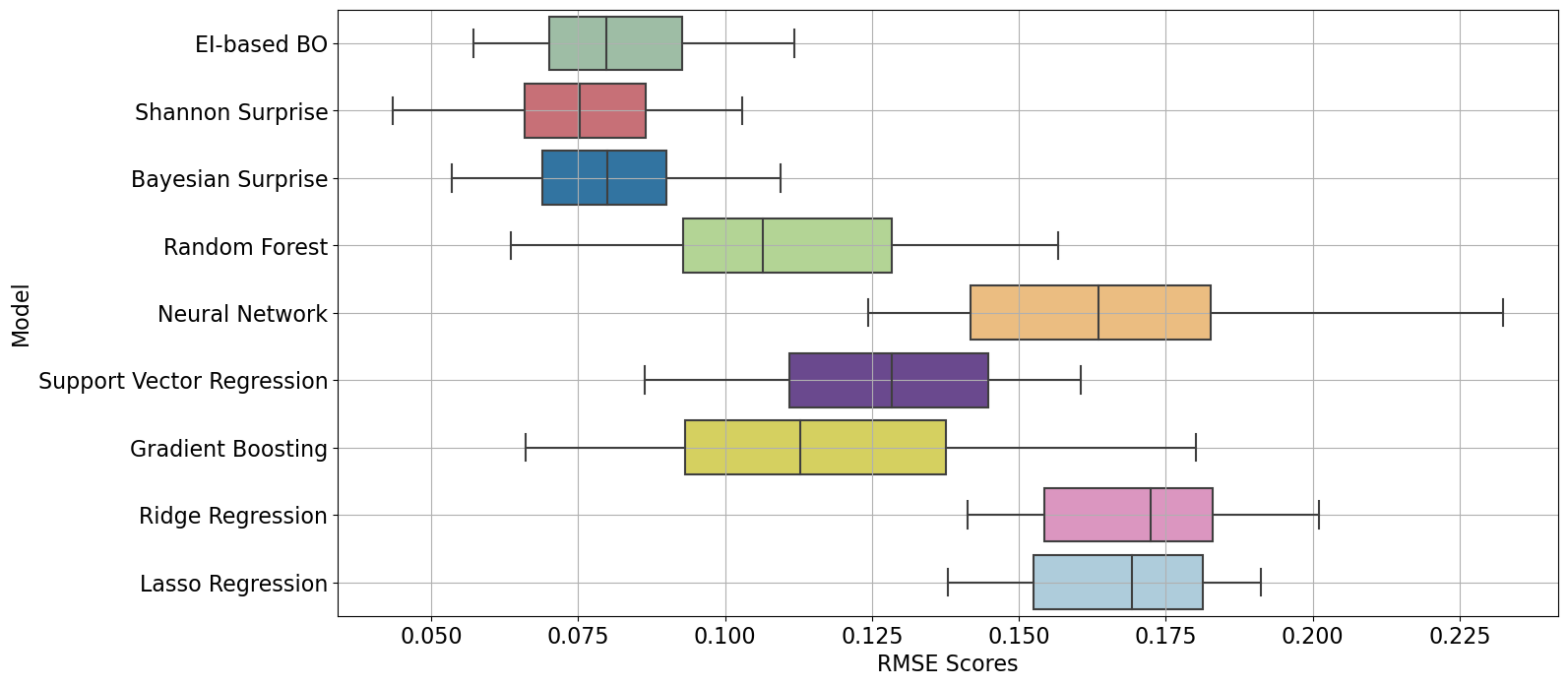}
\caption{Boxplot of RMSE of all the models in predicting the length of melt pool (Scenario II)}
\label{fig:myimage8}
\end{figure*}

The number of experiments, i.e., datapoints available for predicting the melt pool length is less compared to the other two dimensions (melt pool depth and width). Keeping that into consideration, for the three sequential models, we create a set of 30 experiments and a second set of 90 experiments for the initial and sequential experiments, respectively. Similar to the other two predictions, for the six ML models, we have a training set of 200 datapoints in the first scenario and 120 datapoints in the second scenario. The mean and median of these RMSE scores from 20 iterations for both these scenarios are also shown in Table~\ref{tab:rmse_scores_length}. Figure~\ref{fig:myimage8} shows the boxplot of the RMSE scores from 20 iterations for Scenario II for all the models.

\begin{table*}[h]
\centering
\caption{Mean and median of the RMSE scores for all the models in predicting the length}
\label{tab:rmse_scores_length}
\begin{tabular}{|l|c|c|c|c|}
\hline
\multicolumn{1}{|c|}{} & \multicolumn{2}{c|}{\textbf{Scenario I}} & \multicolumn{2}{c|}{\textbf{Scenario II}} \\ \cline{2-5}
\multicolumn{1}{|c|}{\textbf{Model}} & \textbf{Mean} & \textbf{Median} & \textbf{Mean} & \textbf{Median} \\ \hline
EI-based BO & 0.082375 & 0.079799 & 0.082375 & 0.079799 \\ \hline
Shannon Surprise & 0.077949 & 0.075243 & \textbf{0.077949} & \textbf{0.075243} \\ \hline
Bayesian Surprise & 0.078574 & 0.079927 & 0.078574 & 0.079927 \\ \hline
Random Forest & 0.081808 & 0.082618 & 0.109143 & 0.106485 \\ \hline
Neural Network & 0.153148 & 0.151584 & 0.163001 & 0.163548 \\ \hline
Support Vector Regression & 0.103957 & 0.096196 & 0.141573 & 0.128401 \\ \hline
Gradient Boosting & 0.084773 & 0.083035 & 0.116401 & 0.112743 \\ \hline
Ridge Regression & 0.164021 & 0.163434 & 0.172106 & 0.172481 \\ \hline
Lasso Regression & 0.166729 & 0.165621 & 0.167838 & 0.169241 \\ \hline
\end{tabular}
\end{table*}

The results for predicting melt pool length further underscore the effectiveness of our proposed sequential learning approach. Remarkably, in this case, both the Shannon and Bayesian surprise-guided strategies within the SurpriseAF-BO framework excelled, surpassing all ML methods in Scenario II comprehensively. More impressively, these strategies also demonstrated significantly enhanced performance in Scenario I. In Scenario I, the sequential approaches were allocated only 120 data points for their experiments, in contrast to the 200 data points provided to the ML models. Despite this limitation in data availability, the superior performance of the sequential methods underscores the robust predictive capabilities of our sequential learning approach.

In summarizing the results for predicting the dimensions of the melt pool, it is evident that our SurpriseAF-BO approach performs better in environments with limited data and resources. It surpasses both the popular EI-based BO sequential algorithm and various commonly used machine learning algorithms in predictive learning tasks. Even when operating with fewer resources than the ML approaches in Scenario I, our method demonstrates impressive performance. This success can be attributed to its intelligent sample selection mechanism. Unlike the more random or static approach of ML methods, our method strategically chooses each new sample based on the accumulated knowledge from previous data. This guided process of sample selection is a key differentiator for our surprise-based sequential methods, enabling them to outperform traditional ML approaches. By continuously learning and adapting based on each new piece of information, our method efficiently navigates through the data, making the most out of limited resources.

It is important to recognize that in real-world scenarios, sequential experimentation offers significant advantages by directing experimenters on where to conduct their next experiment. However, in the context of our dataset, where the experiments have already been conducted, we adapt this approach to fit a sequential setting. This adaptation involves instructing the acquisition function to choose from only the existing datapoints. This constraint means that, even before any comparison is made, the sequential methods are operating at a disadvantage, limiting their full potential. Despite this limitation, it is noteworthy that these methods still demonstrate superior performance in scenarios with reduced data. 

When comparing the two methods within our approach, the Shannon surprise-guided method consistently shows lower RMSE values than the Bayesian surprise-guided method across all scenarios. This difference is largely due to how each method responds to new data. Shannon surprise quickly identifies and reacts to new, unexpected information, making it highly responsive. On the other hand, Bayesian surprise focuses more on significant changes in data patterns, which means it reacts more slowly, waiting for a larger shift before identifying an observation as a surprise. Therefore, while Shannon surprise tends to recognize a wide range of observations as surprising, Bayesian surprise is more selective, identifying fewer observations as such. This quicker response of the Shannon surprise method contributes to its better performance in our study.

The superior performance of the SurpriseAF-BO approach over the traditional EI-based BO can be attributed to its more nuanced balance between exploration and exploitation, especially in data-scarce environments. The EI acquisition function, commonly used in traditional BO frameworks, has a critical limitation. It tends to prioritize areas offering a small but certain improvement, often overlooking regions that might yield larger gains but with less certainty. This strategy leads to an overemphasis on exploiting local optima, potentially neglecting broader exploration. As a result, while EI is effective for specific function optimization tasks, it can become trapped in local optima due to its limited exploratory behavior.

In contrast, the SurpriseAF-BO approach, with its surprise-based acquisition function, intelligently balances the need to exploit known areas of the landscape for immediate gains and explore new regions for potential significant improvements. This balance is particularly crucial in scenarios with limited data, where each data point's selection must contribute maximally to the model's understanding. The surprise-based approach is adept at identifying and leveraging the most informative data points, thereby approximating the entire landscape more effectively than the EI-based method. This strategic approach to data point selection allows SurpriseAF-BO to excel in approximation tasks, making it a more versatile and effective tool in complex, data-limited environments. 

An interesting finding from our study is that among the six ML techniques we applied, RF and GB exhibited strong performance, even when the number of training samples was halved. Both RF and GB are variants of ensemble learning methods, known for their effectiveness in scenarios with limited data availability. Ensemble methods work by combining predictions from multiple individual models, leading to a more robust overall prediction that is generally less susceptible to overfitting. This characteristic is particularly advantageous in situations where data is scarce. Furthermore, RF and GB are adept at managing irrelevant or noisy data. They achieve this through their inherent feature selection process during tree construction, where they prioritize the most informative features. This ability to focus on relevant features helps minimize the risk of overfitting, making these methods well-suited for resource-constrained scenarios. When comparing these methods to our surprise-based sequential learning approach, it is noteworthy that RF and GB's robustness and feature selection capabilities allow them to remain competitive. However, our surprise-based approach, with its focus on efficiently utilizing limited data and adapting to new information, offers a distinct advantage in dynamically evolving scenarios and in efficiently navigating and learning from complex data landscapes.

\subsection{Phase II: Augmenting SurpiseAF-BO with CTGAN (CT-SurpiseAF-BO)}

In Phase I of our study, we demonstrated the effectiveness of the SurpriseAF-BO algorithm in predicting melt pool geometry, particularly under resource-constrained conditions, where it outperformed traditional ML methods. Building on this success, Phase II aimed to enhance prediction accuracy without incurring additional experimental costs. To achieve this, we integrated a CTGAN into the existing framework of the SurpriseAF-BO algorithm.

The CTGAN model was trained using the same initial data points that were employed for initializing the GP model at the onset of the sequential experiments. Once the CTGAN model was trained, it generated synthetic data, which we then combined with the original dataset. The subsequent steps followed the process outlined earlier in Figure 5. Our evaluation in this phase focused on the modified framework's ability to predict the depth, width, and length of the melt pool. Given our specific interest in enhancing the performance of the surprise-guided sequential learning framework, we concentrated our analysis on the results obtained using the Shannon and Bayesian surprise measures. The hyperparameters used for training the CTGAN model, which played a crucial role in this phase, are detailed in Table 9. This approach allowed us to assess the impact of synthetic data augmentation on the predictive capabilities of our sequential learning framework.

To enhance the quality of the synthetic data, we incorporated prior knowledge into the data generation process, which is a crucial task, particularly in a specialized field like MAM. At first, we performed a careful examination of the process parameters and the target outputs along with their possible combinations and ranges. For example, laser power, scanning velocity, and other parameters have specific operational ranges and physical limits in metal additive manufacturing. Next, we trained the CTGAN with domain-specific knowledge to generate artificial samples that have a similar distribution when compared to the original samples. After the synthetic data were generated following certain conditions using CTGAN, the next step was to check whether the artificial data was realistic and adhered to the operational constraints of the input and output features. For this, knowledge from the existing literature and experts in the field of MAM provided valuable insights that eventually helped identify and remove any unrealistic or implausible data points.

\begin{table*}[h]
\centering
\caption{Description of the hyperparameters selected for the CTGAN model}
\label{tab:hyperparameters_ctgan}
\begin{tabular}{|l|l|c|}
\hline
\textbf{Hyperparameter} & \textbf{Description} & \textbf{Selected Value} \\
\hline
`embedding dim' & Size of the random sample passed to the Generator & 32 \\
\hline
`generator dim` & Size of the output samples for each one of the Residuals & (64, 64) \\
\hline
`discriminator dim' & Size of the output samples for each one of the Discriminator Layers & (64, 64) \\
\hline
`generator lr' & Learning rate for the generator & 0.01 \\
\hline
`discriminator lr' & Learning rate for the discriminator & 0.01 \\
\hline
`batch size' & Number of data samples to process in each step & 25 \\
\hline
`epochs' & Number of training epochs & 500 \\
\hline
`pac' & Number of samples to group together when applying the discriminator & 10 \\
\hline
\end{tabular}
\end{table*}

\subsubsection{Predicting Depth of Melt Pool}

Moving to Phase II, we enhanced the initial dataset used by the SurpriseAF-BO in Phase I, which originally consisted of 50 samples, by adding synthetic data generated through CTGAN. This phase focused on evaluating the performance of the augmented CT-SurpriseAF-BO framework, again using the mean RMSE score. In this phase, we incrementally added 5 synthetic samples each time, while keeping the original set of 50 initial samples (used in Phase I) constant. The outcomes of this approach, demonstrating how the addition of synthetic data influenced the prediction accuracy, are depicted in Figure 9. This visualization provides a clear understanding of the impact of synthetic data augmentation on the predictive capabilities of our sequential learning model.

\begin{figure*}[htbp]
\centering
\includegraphics[width=0.95\linewidth]{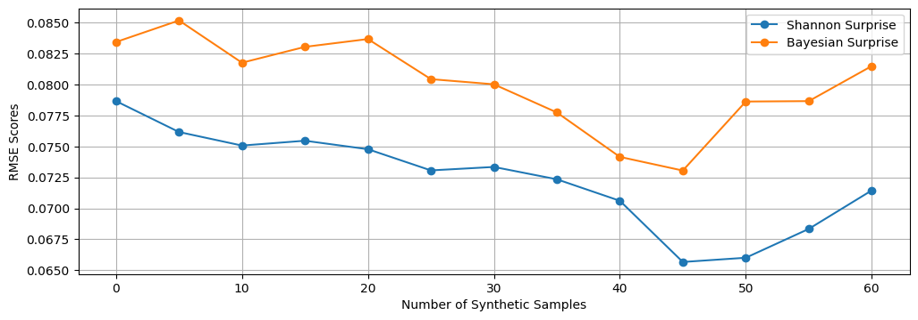}
\caption{RMSE scores of the two surprise guided approaches in Phase II for melt pool depth prediction}
\label{fig:myimage9}
\end{figure*}

Figure~\ref{fig:myimage9} reveals that incorporating synthetic samples into the initial GP model of our proposed framework leads to improved outcomes. The optimal result is observed when 45 synthetic samples, generated using CTGAN, are added. This means the GP model was trained with a total of 95 initial samples, comprising the 50 samples used in Phase I and the 45 new synthetic samples. The number of sequential samples remained consistent with Phase I at 175. This addition of synthetic samples effectively reduced the RMSE scores to 0.0656 for the Shannon surprise-guided approach and 0.0730 for the Bayesian surprise-guided approach. Comparatively, in Phase I, without synthetic samples, the mean RMSE scores for these two approaches were 0.0786 and 0.0834, respectively. Therefore, the inclusion of synthetic samples resulted in an improvement of 16.52\% for the Shannon surprise-guided approach and 12.44\% for the Bayesian surprise-guided approach. These findings highlight the significant impact that synthetic data can have on enhancing the accuracy of sequential learning models.

\subsubsection{Predicting Width of Melt Pool}

In Phase I of our study on predicting melt pool width, we conducted 40 initial experiments and 125 sequential experiments to train the sequential algorithms. For Phase II, we followed a similar approach to that used in predicting melt pool depth, where we augmented our dataset with synthetic data generated from CTGAN. This CTGAN model was trained using the same 40 initial experiments previously utilized. Figure 10 illustrates the impact of varying the number of synthetic samples on the predictive performance, specifically in terms of the RMSE score.

\begin{figure*}[htbp]
\centering
\includegraphics[width=0.95\linewidth]{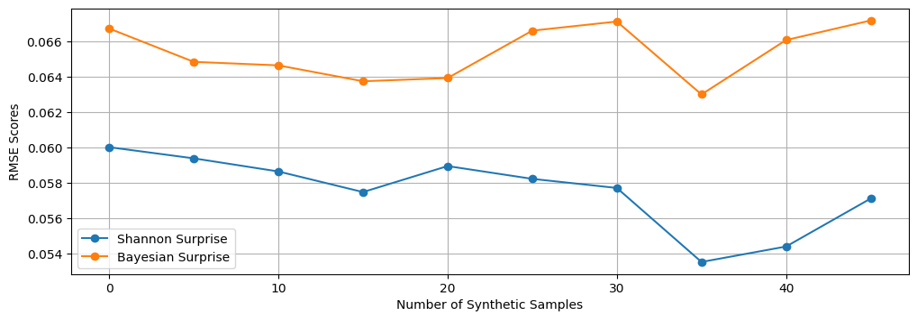}
\caption{RMSE scores of the two surprise guided approaches in Phase II for melt pool width prediction}
\label{fig:myimage10}
\end{figure*}

An analysis of Figure ~\ref{fig:myimage10} reveals that the inclusion of synthetic data generally leads to a decrease in the RMSE score, indicating an improvement in prediction performance. This trend continues up to the addition of about 35 synthetic samples, beyond which the RMSE score begins to increase. Initially, with no synthetic data added, the mean RMSE scores for the Shannon and Bayesian surprise measures were 0.05977 and 0.067024. However, after incorporating synthetic data into the initial model training, the best RMSE scores achieved for these two surprise metrics dropped to 0.0535 and 0.0623, respectively. This improvement translates to a 10.81\% enhancement in prediction performance for the Shannon surprise-guided approach and a 5.58\% improvement for the Bayesian surprise-guided approach. These results underscore the effectiveness of integrating synthetic data in enhancing the accuracy of our predictive models.

\subsubsection{Predicting Length of Melt Pool}

In Phase I of our study, we used 30 initial experiments to establish an initial approximation with a GP model and conducted 90 sequential experiments to predict the length of the melt pool. In Phase II, we continued to use these 30 initial experiments but also incorporated synthetic samples generated using CTGAN. Figure 11 presents the results of this approach. 

\begin{figure*}[htbp]
\centering
\includegraphics[width=0.95\linewidth]{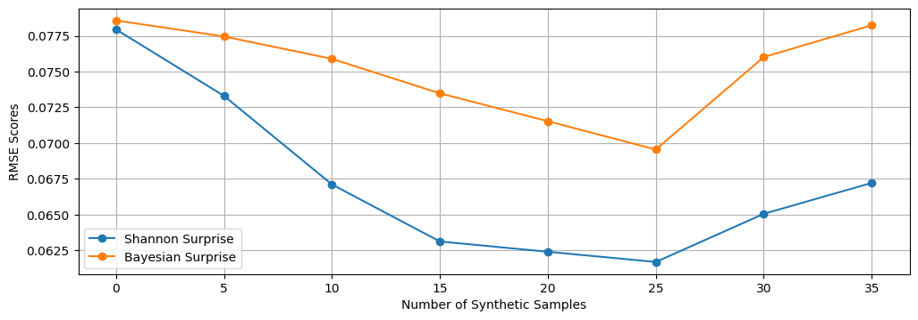}
\caption{RMSE scores of the two surprise guided approaches in Phase II for melt pool length prediction}
\label{fig:myimage11}
\end{figure*}

The analysis of Figure~\ref{fig:myimage11} indicates a clear trend: as we add more synthetic samples to the initial set of 30 experiments, the RMSE scores for both the Shannon and Bayesian surprise measures decrease. The optimal performance is achieved when the dataset includes 25 synthetic samples. At this point, the RMSE scores for the Shannon and Bayesian surprise metrics are 0.0617 and 0.0695, respectively. By integrating CTGAN-generated samples into the initial dataset, we observe a significant improvement in prediction performance: a 20.85\% enhancement for the Shannon surprise measure and an 11.48\% improvement for the Bayesian surprise measure. These results again highlight the effectiveness of using synthetic data to bolster the predictive accuracy of our models.

The analysis of the results reveals an interesting trend in the use of synthetic data for training machine learning models, particularly in the context of predicting melt pool dimensions. Initially, the introduction of synthetic data generated by techniques like CTGAN proves beneficial. It fills gaps in the training dataset, offering a more complete representation of the problem space. This comprehensive dataset enables the GP model to learn more effectively, leading to improved predictions. However, a critical threshold is observed when the addition of synthetic data starts to adversely affect the model's performance. As more artificial samples are added, the RMSE scores begin to increase, indicating a decline in the prediction accuracy for the three melt pool dimensions. This deterioration in performance is commonly attributed to overfitting, a scenario where the GP model, trained on an excessive amount of data, starts to fit too closely to the synthetic data. This overfitting captures noise or irrelevant features rather than the underlying pattern, leading to poorer performance on test or real-world data.

Another factor contributing to this decline is the representativeness and quality of the synthetic data. As the volume of synthetic data increases, there is a risk that it may not accurately represent the true data distribution or may introduce noise, further degrading the model's performance. This is particularly crucial in scenarios where the synthetic data might not perfectly mimic the real-world data's complexities and nuances. Moreover, the results from Phase I highlight the superior capabilities of the proposed sequential learning approach in striking a balance between exploration and exploitation. This balance is key to making informed decisions based on the true underlying data distribution. Excessive reliance on synthetic data can disrupt this balance, leading the GP model to base its decisions more on the synthetic data distribution rather than the actual data distribution. Therefore, while synthetic data can be a powerful tool for enhancing machine learning models, it is crucial to find the right amount that enriches the training dataset without compromising the model's ability to generalize to new, unseen data.

\section{Conclusion}
\label{sec:sample6}

In this research paper, we introduce SurpriseAF-BO and its advanced version, CT-SurpriseAF-BO, as novel sequential learning frameworks designed for the prediction of melt pool geometry in MAM. The study unfolds in two phases, with Phase I demonstrating the superior predictive accuracy of SurpriseAF-BO over six popular machine learning algorithms and the traditional EI-based BO framework in a resource-constrained environment. The standout performance of SurpriseAF-BO is largely attributed to its innovative approach in sample selection, which employs an adaptive feedback-based sequential sampling strategy. This strategy is continuously updated based on the degree of surprise and the existing knowledge of the response surface, proving to be more effective than the random or static sample selection methods typically used in conventional ML that do not account for resource limitations. 

SurpriseAF-BO's effectiveness is further highlighted in its comparison with the traditional EI-based BO. The framework demonstrates a more efficient balance in exploiting known data and exploring new areas. While the EI method often focuses on minor, certain improvements, potentially neglecting larger, uncertain opportunities and thus gravitating towards local maxima, the surprise-based strategies of SurpriseAF-BO adeptly navigate between immediate gains and the exploration of new, potentially advantageous areas. This makes it particularly suitable for approximating the response surface, in contrast to the EI-based BO framework. Between the two surprise measures, the Shannon surprise measure appears to be more effective than its Bayesian counterpart. Being more agile, the Shannon surprise rapidly responds to changes in the underlying response surface. This responsiveness enables it to more effectively guide the optimization algorithm in selecting new experimental locations. In contrast, the Bayesian surprise demands more substantial evidence of changes in data patterns before reacting, resulting in a slower response compared to the Shannon surprise. Phase II of the study builds on the success of phase I by proposing the CT-SurpriseAF-BO framework, which leverages synthetic samples generated by CTGAN to closely mimic real experimental data. This enhancement leads to further improvements in predictive capabilities without the need for additional physical experiments. However, it also reveals a critical threshold beyond which the addition of more synthetic samples can adversely affect performance.

The sequential learning frameworks proposed in this study represent a significant advancement in the manufacturing paradigm, particularly within the realm of MAM. These frameworks offer a dual advantage: firstly, they reduce the necessity for extensive physical experiments by generating accurate predictions with fewer samples, including synthetic data, thereby saving time and reducing costs. Secondly, they enhance predictive accuracy, particularly in determining vital melt pool characteristics, leading to improved control and consistency in MAM processes and, consequently, elevated product quality. While these frameworks are specifically tailored for MAM, the underlying principles of surprise-guided sequential learning and synthetic data augmentation have the potential to transform other manufacturing sectors that require guided physical experiment selection, suggesting a wide-ranging impact on the industry.

 \bibliographystyle{elsarticle-num} 
 \bibliography{cas-refs}





\end{document}